\RequirePackage{booktabs}
\documentclass[pdflatex,sn-mathphys-num]{sn-jnl}
\usepackage{graphicx}%
\usepackage{multirow}%
\usepackage{amsmath,amssymb,amsfonts}%
\usepackage{amsthm}%
\usepackage{mathrsfs}%
\usepackage[title]{appendix}%
\usepackage{xcolor}%
\usepackage{textcomp}%
\usepackage{manyfoot}%
\usepackage{booktabs}%
\usepackage{algorithm}%
\usepackage{algorithmicx}%
\usepackage{algpseudocode}%
\usepackage{listings}%
\usepackage{xstring}
\usepackage{array}
\usepackage{tabularx}
\usepackage{hhline}
\usepackage{graphbox} 
\usepackage{pifont}

\usepackage{dcolumn}
\usepackage{colortbl}
\usepackage{changepage} 
\usepackage{hyperref}
	\hypersetup{colorlinks,	citecolor=blue,	filecolor=blue,	linkcolor=blue,	urlcolor=blue}
\usepackage[margin=0pt,font={small},labelfont={bf}]{caption}
\usepackage{subcaption}
\usepackage{soul}
\usepackage{geometry}
\usepackage{natbib}
\usepackage{titlesec}
	\titleformat{\subsection}{\normalfont\fontsize{12}{14}\bfseries}{\thesubsection}{1em}{}
	\titleformat{\subsubsection}[runin]{\bfseries}{}{}{}[\hspace*{0.8em}]
	\titlespacing*{\section}{0pt}{16pt plus 4pt minus 2pt}{5pt plus 2pt minus 2pt}
	\titlespacing*{\subsection}{0pt}{12pt plus 4pt minus 2pt}{5pt plus 2pt minus 2pt}
    \titlespacing*{\subsubsection}{0pt}{9pt plus 4pt minus 2pt}{2pt plus 2pt minus 2pt}   
\usepackage{makecell}
\usepackage{setspace}
\usepackage{adjustbox}
\usepackage{latexsym}
\usepackage{tikz}
\usepackage[most]{tcolorbox}
\usepackage{enumitem}
\makeatletter
\newcommand\footnoteref[1]{\protected@xdef\@thefnmark{\ref{#1}}\@footnotemark}
\makeatother
\def\hd{HD}
\def\hdlong{Hausdorff distance}
\def\Hdlong{\hdlong}
\def\hdpercp{HD$_{\!p}$} 
\newcommand{\hdpercn}[1]{HD\textsubscript{#1}} 

\def\masd{MASD}
\def\masdlong{mean average surface distance}
\def\Masdlong{Mean average surface distance}
\def\assd{ASSD}
\def\assdlong{average symmetric surface distance}
\def\Assdlong{Average symmetric surface distance}
\def\nsd{NSD}
\def\nsdlong{normalized surface distance}
\def\Nsdlong{Normalized surface distance}
\def\biou{BIoU}
\def\bioulong{boundary intersection over union}
\def\Bioulong{Boundary intersection over union}
\def\dsc{DSC}
\def\dsclong{Dice similarity coefficient}
\def\Dsclong{\dsclong}
\def\iou{IoU}
\def\ioulong{intersection over union}

\newcommand{\thspace}{\hspace{0.2em}}
\def\googledeepmind{Google\thspace DeepMind}
\def\googledeepmindtt{\texttt{\small\googledeepmind}}
\def\metricsreloaded{Metrics\thspace Reloaded}
\def\metricsreloadedtt{\texttt{\small\metricsreloaded}}
\def\metricsreloadedit{\textit{\metricsreloaded}}
\def\meshmetrics{MeshMetrics}
\def\meshmetricstt{\texttt{\small\meshmetrics}}

\def\segA{$A$}

\def\segB{$B$}

\def\maccuDab{\hat{d}_{AB}}
\def\maccuDba{\hat{d}_{BA}}
\def\mpercDab{d_{AB}^{\,(p)}}
\def\mpercDba{d_{BA}^{\,(p)}}
\def\mtolpmA{\partial A^{(\pm\tau)}}
\def\mtolpmB{\partial B^{\,(\pm\tau)}}

\def\mbufferA{\mathcal{N}_{\partial A}^{\,(-\tau)}}
\def\mbufferB{\mathcal{N}_{\partial B}^{\,(-\tau)}}

\def\msizeA{|\partial A|}
\def\msizeB{|\partial B|}
\def\mdistAB{d_{a, \partial B}}
\def\mdistBA{d_{b, \partial A}}
\def\mDistSetAB{D_{\!A\!B}}
\def\mDistSetBA{D_{\!B\!A}}
\def\mBoundarySetA{\Delta_{\!A}}
\def\mBoundarySetB{\Delta_{\!B}}
\def\mdistAx{d_{x, \partial A}}
\def\mdistBx{d_{x, \partial B}}

\def\spacingOneOne{$(1.0{,}\,1.0)$\,mm}
\def\spacingPointZeroSevenPointZeroSeven{$(0.07{,}\,0.07)$\,mm}
\def\spacingPointZeroSevenOne{$(0.07{,}\,1.0)$\,mm}

\def\spacingOneOneOne{$(1.0{,}\,1.0{,}\,1.0)$\,mm}

\def\spacingTwoTwoTwo{$(2.0{,}\,2.0{,}\,2.0)$\,mm}

\def\spacingHalfHalfTwo{$(0.5{,}\,0.5{,}\,2.0)$\,mm}

\def\Nan{\texttt{NaN}} 
\def\Inf{\texttt{$\infty$}} 

\newcommand{\hrefx}[1]{\href{#1}{#1}}
\begin{document}
\title[MeshMetrics: A Precise Implementation of Distance-Based Image Segmentation Metrics]{\centering \textbf{{MeshMetrics}: A Precise Implementation of Distance-Based Image Segmentation Metrics}}
\author[*]{\fnm{Ga\v{s}per} \sur{Podobnik}}
\author{\fnm{Toma\v{z}} \sur{Vrtovec}}
\affil{\normalsize
	\orgname{University of Ljubljana},
	\orgdiv{Faculty of Electrical Engineering},  \orgaddress{\street{Tr\v{z}a\v{s}ka cesta 25}, \postcode{SI-1000} \city{Ljubljana}, \country{Slovenia}}}
\abstract{%
The surge of research in image segmentation has yielded remarkable performance gains but also exposed a reproducibility crisis. A major contributor is performance evaluation, where both \textit{selection} and \textit{implementation} of metrics play critical roles. While recent efforts have improved the former, the reliability of metric implementation has received far less attention. Pitfalls in distance-based metric implementation can lead to considerable discrepancies between common open-source tools, for instance, exceeding 100\,mm for the Hausdorff distance and 30\%pt for the normalized surface distance for the same pair of segmentations. To address these pitfalls, we introduce \meshmetricstt, a \textit{mesh-based} framework that provides a more precise computation of distance-based metrics than conventional \textit{grid-based} approaches. Through theoretical analysis and empirical validation, we demonstrate that \meshmetricstt\ achieves higher accuracy and precision than established tools, and is substantially less affected by discretization artifacts, such as distance quantization. We release \meshmetricstt\ as an open-source Python package, available at \hrefx{https://github.com/gasperpodobnik/MeshMetrics}.}
\maketitle
\section{Introduction}
\label{sec:introduction}
\begin{figure}[!t]
    \centering
    \includegraphics[width=0.50\textwidth]{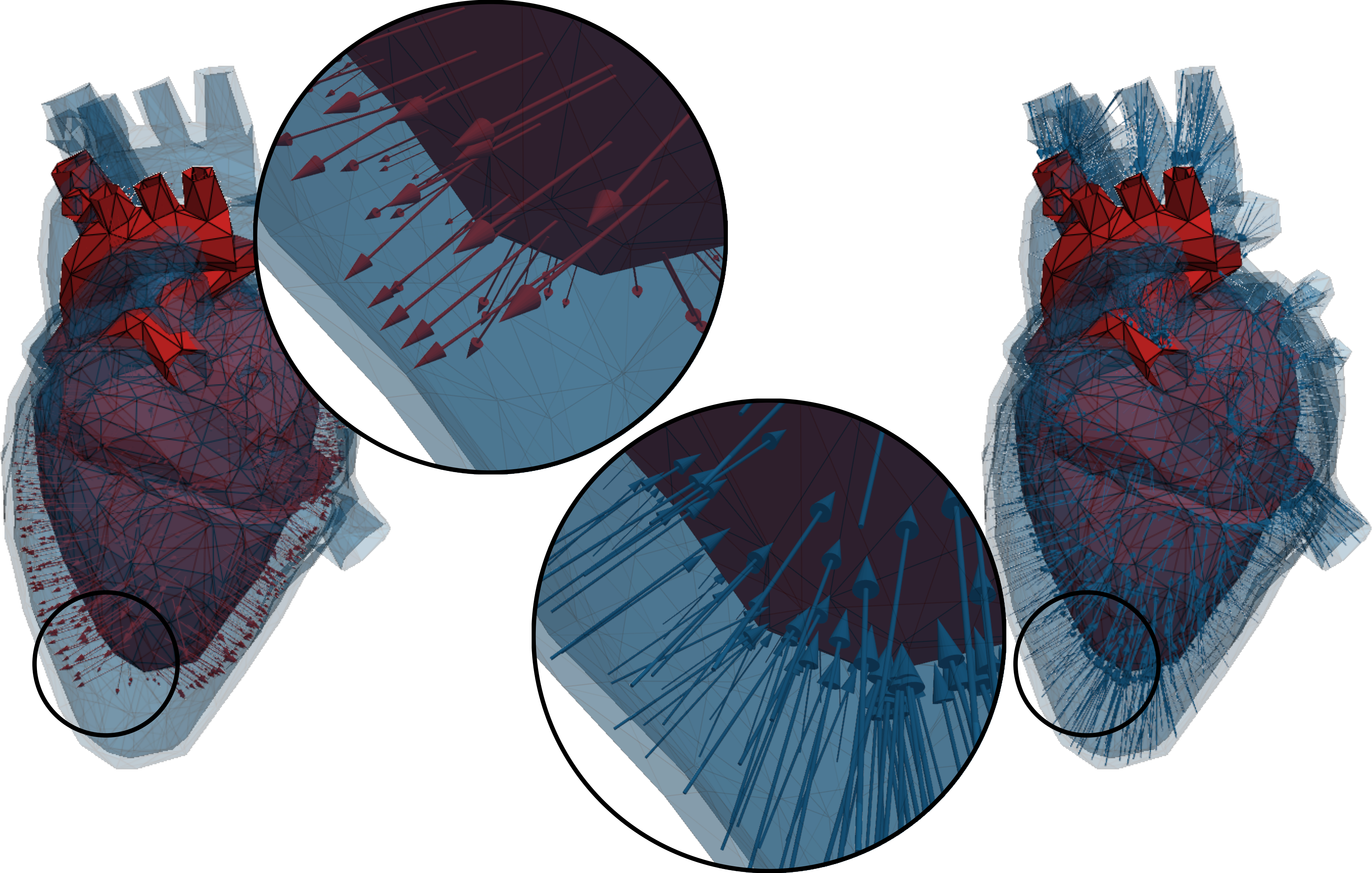}
    \caption{A schematic illustration of distance computation between two segmentations using a mesh-based representation of the heart.}
    \label{fig:sketch}
    \vspace{-3ex}
\end{figure}

Image segmentation is a fundamental task in computer vision that aims to partition an image into pixels or voxels corresponding to structural regions of interest. Advances in computational resources and algorithmic design have fueled the development of highly accurate automatic segmentation methods~\cite{Shaker2024unetrpp, Wang2025ContourAware, Shen2025DeepLearning}. Their performance is typically assessed using geometric metrics, which provide fast and objective quantitative evaluations of segmentation quality. The most widely used metrics include overlap-based, such as the \textit{\Dsclong} (\dsc)~\cite{Dice1945Measures} and \textit{\ioulong} (\iou)~\cite{Jaccard1912Distribution}, and distance-based (Fig.~\ref{fig:sketch}), such as the \textit{\Hdlong} (\hd)~\cite{Hausdorff1914-SetTheory} with its $p$-th percentile variants (\hdpercp)~\cite{Huttenlocher1993-HD-Algorithm}, \textit{\masdlong} (\masd)~\cite{Sluimer2005-MASD} and \textit{\assdlong} (\assd)~\cite{Lamecker2004-ASSD}, as well as the recently proposed \textit{\nsdlong} (\nsd)~\cite{Nikolov2021-DeepMind} and \textit{\bioulong} (\biou)~\cite{Cheng2021BoundaryIoU}.
\par
The increasing variety of segmentation metrics has improved validation of automatic segmentation but also introduced several challenges: selecting representative metrics, enabling fair benchmarking, addressing correlations, ensuring consistent nomenclature, and, critically, providing correct implementations. While issues related to metric selection have been widely acknowledged and addressed through community-driven guidelines~\cite{Taha2015-EvaluateSegmentation, Muller2022-MISeval, Maier-Hein2024-Metrics-Reloaded, DrukkerMIDRC2024}, the correctness and consistency of implementations have received considerably less attention. Not all metrics are equally affected: while overlap-based metrics can be robustly implemented in the grid domain (i.e.\ on segmentation masks), distance-based metrics are more nuanced, since translating mathematical definitions into discretized calculations requires multiple non-trivial computational steps~\cite{Reinke2024-Metrics-Pitfalls, Podobnik2024HDilemma}.
\subsection{Related work}
Issues of incorrect metric \textit{implementation} have been noted in studies primarily concerned with metric \textit{selection}~\cite{Muller2022-MISeval, Reinke2024-Metrics-Pitfalls}, but they have never been systematically scrutinized in terms of their practical implementation. Moreover, the lack of awareness of these issues is evident from the frequent omission of the implementation source (i.e.\ open- or closed-source tools) in publications, as well as the inconsistent reporting of software versions, thus implicitly assuming that all implementations are equivalent. We began our investigation after obtaining considerably different \hdpercn{95} values using two open-source tools applied to the same pair of segmentations, which prompted a systematic study of \hd\ and \hdpercp\ across five open-source tools~\cite{Podobnik2024HDilemma} that resulted in \hdpercn{95} discrepancies exceeding $100$\,mm. In a subsequent extended study~\cite{Podobnik2024-UnderstandingPitfalls-arxiv}, we broadened our analysis to $11$ open-source tools and five distance-based metrics (\hdpercp, \masd, \assd, \nsd, \biou)\footnote{This set corresponds to the distance-based metrics included in the widely adopted \metricsreloadedit\ guidelines for metric selection~\cite{Maier-Hein2024-Metrics-Reloaded}.}, which resulted in a comprehensive overview of implementation pitfalls. We found that the open-source tools differ in the (i)~\textit{boundary extraction algorithm} (five distinct algorithms are used), (ii)~\textit{mathematical definitions} (multiple and non-equivalent definitions were identified), and~(iii) \textit{edge-case handling} (inconsistent results when one or both input segmentations are empty). These pitfalls manifested differently in two and three dimensions (2D and 3D), and also varied with pixel/voxel size, with the largest absolute discrepancies exceeding $100$\,mm, $40$\,mm, $20$\,mm, $30$\%pt, and $30$\%pt for \hdpercp, \masd, \assd, \nsd, and \biou, respectively. While some discrepancies stemmed from ambiguous mathematical definitions that can be straightforwardly resolved by fixing the code, other reflected fundamentally different boundary extraction strategies and computational paradigms. Notably, all evaluated open-source tools computed distance-based metrics exclusively in the grid domain, making them inherently susceptible to discretization artifacts.
\subsection{Motivation}
In this study, we address these pitfalls by unifying mathematical definitions and by translating metric computation from the grid to the mesh domain. The \textit{mesh-based} paradigm enables a more accurate boundary extraction, a more precise distance computation (i.e.\ point-to-surface rather than point-to-point, cf.\ Fig.~\ref{fig:sketch}), and explicit weighting by boundary element sizes that exactly approximates surface integrals, which cannot be achieved in the grid domain. We release the resulting implementation as \meshmetricstt, an open-source Python package available at \hrefx{https://github.com/gasperpodobnik/MeshMetrics}.
\begin{figure*}[!t]
    \centering

\def\hdpercp{HD$_{p}$} 
\def\masd{MASD}
\def\assd{ASSD}
\def\nsd{NSD}
\def\nsdtau{NSD$_{\tau}$}
\def\biou{BIoU}
\def\bioutau{BIoU$_{\tau}$}
\def\maccuDab{\hat{d}_{AB}}
\def\maccuDba{\hat{d}_{BA}}
\def\mpercDab{d_{AB}^{\,(p)}}
\def\mpercDba{d_{BA}^{\,(p)}}
\def\mtolpmA{\partial A^{(\pm\tau)}}
\def\mtolpmB{\partial B^{\,(\pm\tau)}}

\def\mbufferA{\mathcal{N}_{\partial A}^{\,(-\tau)}}
\def\mbufferB{\mathcal{N}_{\partial B}^{\,(-\tau)}}

\def\msizeA{|\partial A|}
\def\msizeB{|\partial B|}
\def\mdistAB{d_{a, \partial B}}
\def\mdistBA{d_{b, \partial A}}
\def\mdistAx{d_{x, \partial A}}
\def\mdistBx{d_{x, \partial B}}

\def\D{\mathrm{d}} 

\newcolumntype{M}[1]{>{\centering\arraybackslash}m{#1}}
\newcolumntype{C}{>{\centering\arraybackslash}X}
\newcolumntype{x}[1]{>{\arraybackslash\hspace{0pt}}p{#1}}

\definecolor{TMItitlecolor}{rgb}{0.0,0.263,0.576}
\definecolor{TMIsubsectioncolor}{rgb}{0.0,0.541,0.855}

\vspace{2ex}
    \centering
    \begin{tcolorbox}[
    colback=gray!10!white, 
    colframe=black!80!white, 
    fonttitle=\bfseries, 
    colbacktitle=TMItitlecolor,
    halign title=center,
    boxrule=0.5pt, 
    arc=3pt, 
    left=2pt, 
    right=2pt, 
    top=2pt, 
    bottom=0pt]
    {\large
    \begin{tabularx}{\textwidth}{C C C}
    \noalign{\smallskip}
    Continuous domain & Discrete domain (mesh) & Discrete domain (grid)
    \end{tabularx}
    }
    \begin{tikzpicture}
        \node[anchor=south west,inner sep=0] (image) at (0,0) {\includegraphics[width=\linewidth]{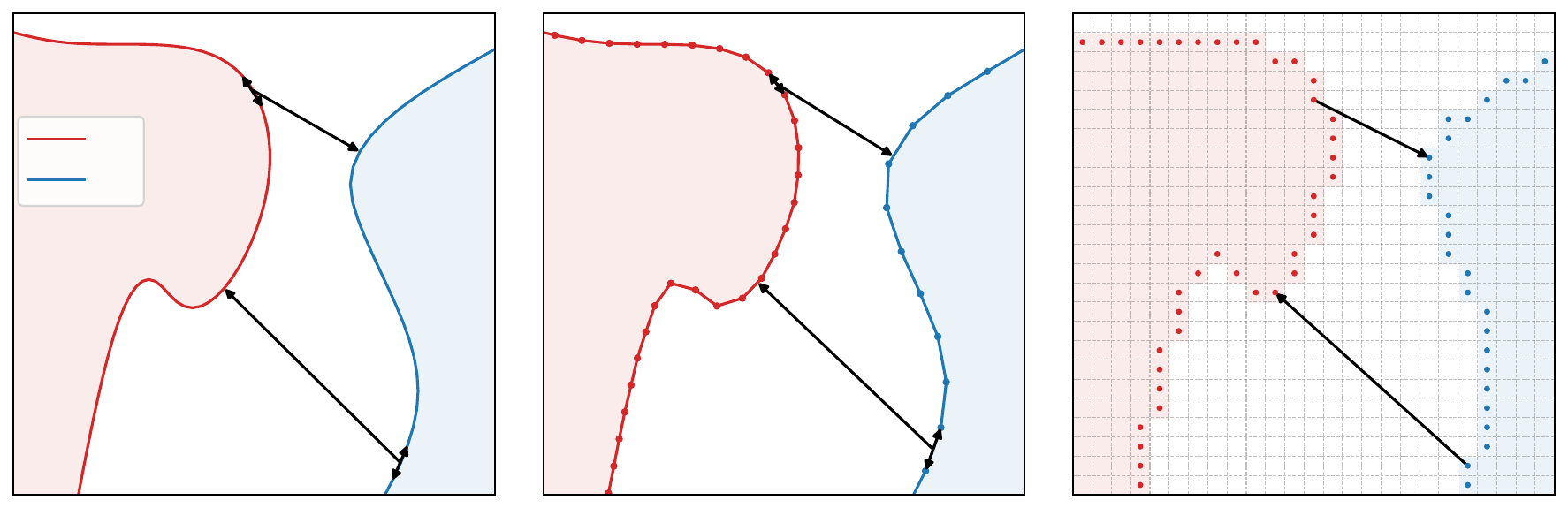}};
        \begin{scope}[x={(image.south east)},y={(image.north west)}]
            \node[fill=none, align=left, anchor=west] at (0.055,0.73) {$\partial A$};
            \node[fill=none, align=left, anchor=west] at (0.055,0.65) {$\partial B$};
            \node[fill=none, align=left, anchor=west] at (0.02,0.4) {$A$};
            \node[fill=none, align=left, anchor=west] at (0.28,0.4) {$B$};
            \node[fill=none, align=left, anchor=west] at (0.128,0.81) {${\D a}$};
            \node[fill=none, align=left, anchor=west] at (0.252,0.09) {${\D b}$};
            \node[fill=none, rotate=-29.5] at (0.20,0.788) {$\mdistAB$};
            \node[fill=none, rotate=-43] at (0.21,0.27) {$\mdistBA$};
            \draw[->, line width=1.5pt] (0.17, 0.0) -- ++(0,-0.05);
            \node[fill=none, align=left, anchor=west] at (0.35,0.4) {$A$};
            \node[fill=none, align=left, anchor=west] at (0.615,0.4) {$B$};
            \node[fill=none, align=left, anchor=west] at (0.46,0.82) {${\delta a}$};
            \node[fill=none, align=left, anchor=west] at (0.594,0.11) {${\delta b}$};
            \node[fill=none, rotate=-32] at (0.54,0.784) {$\mdistAB$};
            \node[fill=none, rotate=-43] at (0.548,0.295) {$\mdistBA$};
            \draw[->, line width=1.5pt] (0.5, 0.0) -- ++(0,-0.05);
            \node[fill=none, align=left, anchor=west] at (0.69,0.4) {$A$};
            \node[fill=none, align=left, anchor=west] at (0.955,0.4) {$B$};
            \node[fill=none, align=left, anchor=west] at (0.805,0.92) {$a\,\in\,\mathcal{A}$};
            \node[fill=none, align=left, anchor=west] at (0.92,0.885) {$b\,\in\,\mathcal{B}$};
            \node[fill=none, rotate=-26.5] at (0.88,0.771) {$\mdistAB$};
            \node[fill=none, rotate=-42.5] at (0.88,0.28) {$\mdistBA$};
            \draw[->, line width=1.5pt] (0.83, 0.0) -- ++(0,-0.05);
            \node[fill=none, align=center, anchor=center] at (0.331,0.5) {\huge $\approx$};
            \node[fill=none, align=center, anchor=center] at (0.669,0.5) {\huge $\neq$};
        \end{scope}
    \end{tikzpicture}
    \vspace{-18pt}
    \begin{tcolorbox}[
    colback=gray!10!white, 
    colframe=black!80!white, 
    title=(a) Accumulated directed distance $\maccuDab$,
    fonttitle=\bfseries,     
    colbacktitle=TMIsubsectioncolor,
    opacitybacktitle=0.9,
    halign title=center,
    boxrule=0.5pt, 
    arc=3pt, 
    left=0pt, 
    right=0pt, 
    top=-20pt, 
    bottom=-18pt]
    \begin{tabularx}{\textwidth}{C C C}
    $$\oint_{\partial A} \!\!\! \mdistAB \, {\D a}$$ & 
    $$\sum_{\partial A} \mdistAB \, \delta a$$ & 
    $$\sum_{\partial A} \mdistAB$$\\
    \end{tabularx}
    \end{tcolorbox}
    \begin{tcolorbox}[
    colback=gray!10!white, 
    colframe=black!80!white, 
    title=(b) $\mathbf{p}$-th percentile directed distance $\mpercDab$,
    fonttitle=\bfseries,     
    colbacktitle=TMIsubsectioncolor,
    opacitybacktitle=0.9,
    halign title=center,
    boxrule=0.5pt, 
    arc=3pt, 
    left=0pt, 
    right=0pt, 
    top=-8pt, 
    bottom=-8pt]
    {\small
    \begin{tabularx}{\textwidth}{C C C}
    $$\inf \Bigg\{\! t \!\in\! \mathbb{R} \bigg| \frac{1}{\msizeA} \! \int_{\partial A} \!\!\!\! \mathbb{I}\Big[\mdistAB \leq t\Big] \, \D a \geq p \!\Bigg\}$$ & 
    $$\inf \Bigg\{\! t \!\in\! \mathbb{R} \bigg| \frac{1}{\msizeA} \! \sum_{\partial A} \mathbb{I}\Big[\mdistAB \leq t\Big] \, \delta a \geq p \!\Bigg\}$$ & 
    $$\inf \Bigg\{\! t \!\in\! \mathbb{R} \bigg| \frac{1}{\msizeA} \! \sum_{\partial A} \mathbb{I}\Big[\mdistAB \leq t\Big] \geq p \!\Bigg\}$$ \\
    \end{tabularx}
    }
    \end{tcolorbox}
    \begin{tcolorbox}[
    colback=gray!10!white, 
    colframe=black!80!white, 
    title=(c) Boundary within tolerance $\mtolpmA$,
    fonttitle=\bfseries,     
    colbacktitle=TMIsubsectioncolor,
    opacitybacktitle=0.9,
    halign title=center,
    boxrule=0.5pt, 
    arc=3pt, 
    left=0pt, 
    right=0pt, 
    top=-20pt, 
    bottom=-18pt]
    \begin{tabularx}{\textwidth}{C C C}
    $$\int_{\partial A} \! \mathbb{I}\Big[\,|\,\mdistAB\,| \leq \tau \,\Big] \, \D a $$ & 
    $$\sum_{\partial A}    \mathbb{I}\Big[\,|\,\mdistAB\,| \leq \tau \,\Big] \, \delta a $$ & 
    $$\sum_{\partial A}    \mathbb{I}\Big[\,|\,\mdistAB\,| \leq \tau \,\Big] $$ \\
    \end{tabularx}
    \end{tcolorbox}
    \begin{tcolorbox}[
    colback=gray!10!white, 
    colframe=black!80!white, 
    title=(d) Interior neighborhood of the boundary $\mbufferA$,
    fonttitle=\bfseries,     
    colbacktitle=TMIsubsectioncolor,
    opacitybacktitle=0.9,
    halign title=center,
    boxrule=0.5pt, 
    arc=3pt, 
    left=0pt, 
    right=0pt, 
    top=-18pt, 
    bottom=-18pt]
    \begin{tabularx}{\textwidth}{C C C}
    $$\left\{ x \in \mathbb{R}^{(n)} \middle| -\tau < \mdistAx < 0 \right\}$$ & 
    $$\left\{ x \in \mathbb{R}^{(n)} \middle| -\tau < \phi_{M_{\partial A}}(x) < 0 \middle. \right\}$$ & 
    $$\left\{ x \in \mathcal{G}_h^{(n)} \middle| -\tau < \mdistAx < 0 \right\}$$ \\
    \end{tabularx}
    \end{tcolorbox}
    \begin{tcolorbox}[
    colback=gray!10!white, 
    colframe=black!80!white, 
    boxrule=0.5pt, 
    arc=3pt, 
    left=6pt, 
    right=6pt, 
    top=5pt, 
    bottom=-1pt]
    \begin{tabularx}{\textwidth}{C C C}
    $\msizeA = \oint_{\partial A} \! {\D a}$ & 
    $\msizeA = \sum_{\partial A} \! {\delta a}$ & 
    $\msizeA = |\mathcal{A}|$ (cardinality of $\mathcal{A}$) \\
    \multicolumn{3}{p{0.98\textwidth}}{
    \rule{\linewidth}{0.4pt}
    Let $A$ and $B$ denote two segmentations, and $\partial A$ and $\partial B$ their respective boundaries.
    \begin{itemize}[noitemsep, topsep=0pt]
        \item In the \textit{continuous} domain, the boundary is smooth, and distances are integrated over infinitesimal boundary elements, denoted as $\D a$, along $\partial A$ (or $\D b$ along $\partial B$).
        \item In the \textit{discrete mesh-based} domain, the boundary consist of line segments/surface triangles in 2D/3D, denoted as $\delta a$ (or $\delta b$), and distances are approximated by a weighted sum over these boundary elements.
        \item In the \textit{discrete grid-based} domain, the boundary is represented by a finite set of boundary points $\mathcal{A}$ (or $\mathcal{B}$), and distances are computed as an average over these points.
    \end{itemize}
    \vspace{-0.7em}
    \rule{\linewidth}{0.4pt}
    \parbox{0.97\textwidth}{\vspace*{0.3em}\small\setstretch{1.1}\textit{Notations:} 
    $\mdistAB = \inf_{b\,\in\,\partial B} \, \lVert a - b\rVert_2$ -- the Euclidean distance from a point on $\partial A$ to $\partial B$ (and vice-versa for $\mdistBA$); 
    $\mathbb{I}[\cdot]=1\text{ if true},\,0\text{ otherwise}$ -- an indicator function;
    $\phi_{M_{\partial A}}(x)$ -- a signed distance function to the mesh boundary, negative inside $A$ and positive outside;
    $\mathcal{G}_h^{\smash{(n)}} = \big\{\, (i_1 h_1, \dots, i_n h_n) \ \big|\ i_k \in \mathbb{Z}, 1 \le k \le n \big\}$ -- a set of all regularly spaced points (grid) in $n$-dimensional space ($n \in \{2,3\}$), where $h_k$ is the pixel/voxel size along $k$-th coordinate direction.}
    }
    \end{tabularx}
    \end{tcolorbox}
    \begin{tcolorbox}[
    colback=gray!10!white, 
    colframe=black!80!white, 
    title=(e) Definitions of distance-based metrics,
    fonttitle=\bfseries,     
    colbacktitle=TMItitlecolor,
    opacitybacktitle=0.9,
    halign title=center,
    boxrule=0.5pt, 
    arc=3pt, 
    left=6pt, 
    right=6pt, 
    top=0pt, 
    bottom=0pt]
    \begin{tabularx}{\textwidth}{C C C C C}
    \hdpercp\ & \masd\ & \assd\ & \nsdtau\ & \bioutau \\
    \noalign{\smallskip} \hline \noalign{\smallskip}
    $\max\left(\mpercDab, \mpercDba\right)$ &
    $\dfrac{1}{2} \left( \dfrac{\maccuDab}{\msizeA} + \dfrac{\maccuDba}{\msizeB} \right)$ &
    $\dfrac{\maccuDab + \maccuDba}{\msizeA + \msizeB}$ &
    $\dfrac{\mtolpmA + \mtolpmB}{\msizeA + \msizeB}$ &
    $\dfrac{|\,\mbufferA \cap \mbufferB|}{|\,\mbufferA \cup \mbufferB|}$ \\
    \end{tabularx}
    \end{tcolorbox}
    \end{tcolorbox}

    \caption{Differences among continuous, discrete mesh-based, and discrete grid-based computational paradigms of distance-based metrics.}
    \label{fig:computation-paradigms}
\end{figure*}
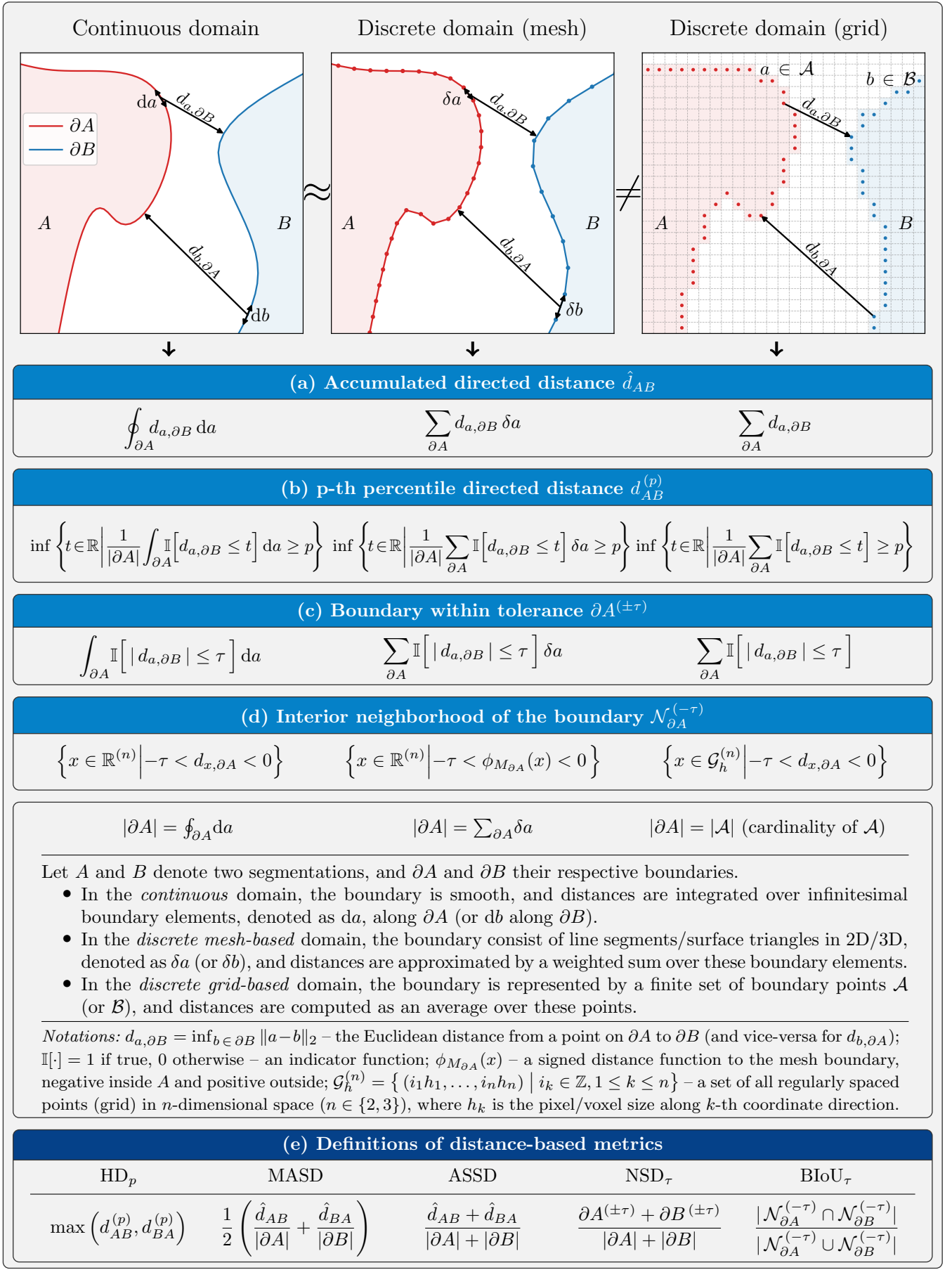
\begin{figure*}[!t]
    \centering
    \centerline{\includegraphics[width=\textwidth]{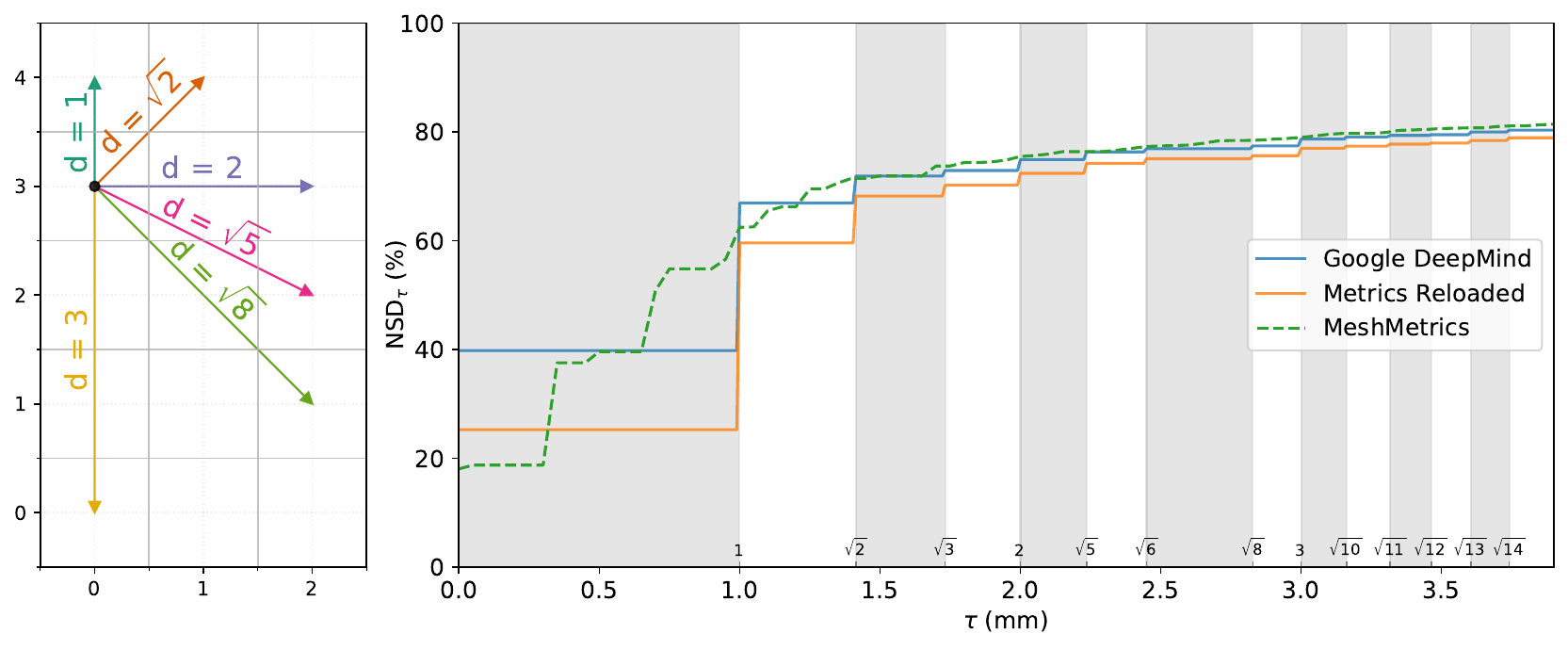}}
    \vspace{-1ex}
    \caption{A schematic illustration of distance quantization in 2D (\textit{left}) and an empirical example of the relationship between the distance tolerance $\tau$ and \nsd\ (with quantized distances annotated in gray) for a 3D case with unit voxel size (\textit{right}).}
    \label{fig:quantized-distances}
    \vspace{-1ex}
\end{figure*}
\section{Methods}\label{sec:Methods}
We first present the paradigms for calculating distance-based metrics and provide conceptual arguments for the calculation in the mesh domain. We then describe the experimental setup used to empirically evaluate \meshmetricstt\ against two established open-source tools. 
\subsection{Calculation paradigms}
Segmentations\footnote{Without loss of generality, we assume binary segmentations, however, the same conclusions can be applied to multilabel segmentations.} can be defined as continuous regions with clearly-defined boundaries (Fig.~\ref{fig:computation-paradigms}, \textit{top left}), represented as explicit surfaces (meshes), implicit surfaces (level sets or radial basis functions)~\cite{Gibou2018Level-set}, volumes (binary masks), parametric surfaces (splines, NURBS)~\cite{Segars2010XCAT}, or point clouds\footnote{However, point clouds are inherently a lossy representation of segmentations, as there is no unique mapping to the corresponding boundary.}. Among these representations, the volumetric segmentation mask (Fig.~\ref{fig:computation-paradigms}, \textit{top right}) is most commonly used in computer vision, particularly in medical image analysis, as images are typically defined on regular grids. However, grid-based representations introduce discretization artifacts because they are inherently limited by the image element size (i.e.\ pixel/voxel size in 2D/3D). These artifacts can be mitigated by increasing the grid density, but this leads to a quadratic/cubic increase in the number of pixels/voxels, which is computationally inefficient. Furthermore, the segmentation boundary is not explicitly encoded in the grid domain, making distance-based metric calculation less convenient, as it requires a preprocessing step to extract the boundary from the segmentation mask~\cite{Podobnik2024HDilemma}. By contrast, mesh representations offer a more flexible description of segmentations, as they explicitly encode and can, in principle, approximate the segmentation boundary to an arbitrary degree of detail, achieving in practice a reasonable accuracy without dramatically increasing the number of boundary elements (Fig.~\ref{fig:sketch} and Fig.~\ref{fig:computation-paradigms}, \textit{top center}). Crucially, the choice of representation has direct implications for metric calculation. As Fig.~\ref{fig:computation-paradigms} illustrates, analytical calculations in the continuous domain can be naturally approximated with a mesh representation, but not with a grid representation, as explained below.
\subsubsection{Implications for \masd, \assd, and HD$_{p}$.}
Let $A$ and $B$ denote two segmentations, and $\partial A$ and $\partial B$ their respective boundaries. The \textit{accumulated directed distance} $\maccuDab$ (Fig.~\ref{fig:computation-paradigms}a), which represents an intermediate step in \masd\ and \assd\ calculation, is analytically defined as the closed integral of the distances over boundary $\partial A$ to the opposing boundary $\partial B$. In the mesh-based formulation, boundaries are approximated by line segments/surface elements in 2D/3D, enabling a direct discretization of the integral as a summation. Conversely, the grid domain does not provide a clear concept of the boundary, as all boundary extraction algorithms essentially return a point cloud without the notion of boundary elements (cf.\ continuous boundary in Fig.~\ref{fig:computation-paradigms}, \textit{top center}, vs.\ non-connected dots in Fig.~\ref{fig:computation-paradigms}, \textit{top right})\footnote{This issue is further affected by the large variability of boundary extraction algorithms across open-source tools~\cite{Podobnik2024-UnderstandingPitfalls-arxiv}.}. In the grid domain, accumulated directed distances are often computed by summing the distances to the opposing point cloud (i.e.\ point-to-point distances vs.\ point-to-surface distance in the mesh domain). This implicitly assumes that points are uniformly distributed along the boundary, which is generally false, as points on flatter regions cover a smaller effective boundary than points on the corners. While these inaccuracies may seem minor, they can substantially impact metric values when aggregated, as evidenced by the large resulting deviations~\cite{Podobnik2024-UnderstandingPitfalls-arxiv}. The same pitfall arises for percentile-based metrics such as \hdpercp, formally defined as the minimal distance $t$ so that at least $p$ percent of $\partial A$ lies within distance $t$ of $\partial B$. Importantly, this definition is based on the \textit{boundary length/area} lying within $t$, whereas grid-based implementations typically oversimplify it to the $p$-th percentile of the set of distances\footnote{Here, the $p$-th percentile is computed by sorting all values in the set and selecting the element at the corresponding $p$-th rank determined by the cardinality of that set.}, implicitly assuming a uniform sampling of boundary points.
\subsubsection{Implications for \nsd.}
While the aforementioned pitfalls also affect the calculation of \nsd, this metric is additionally affected by distance quantization. The \textit{boundary within tolerance} $\mtolpmA$ (Fig.~\ref{fig:computation-paradigms}c) represents the proportion of the boundary lying within a predefined distance tolerance $\tau$, making the choice of $\tau$ critical~\cite{Nikolov2021-DeepMind}. In medical imaging, $\tau$ is typically set between 1--3\,mm, therefore often comparable to the pixel/voxel size of computed tomography (CT) and magnetic resonance (MR) images. In the grid domain, however, only a limited set of distances can be computed due to the discrete arrangement of points according to pixel/voxel size (Fig.~\ref{fig:quantized-distances}, \textit{left}), causing  \nsd\ to jump in discrete steps rather than increase smoothly with $\tau$ (Fig.~\ref{fig:quantized-distances}, \textit{right}). By contrast, in the mesh domain, distances are computed in a point-to-surface manner rather than point-to-point, which improves precision and better mitigates the effect of quantized distances.
\subsubsection{Implications for \biou.}
\biou\ is a hybrid between overlap and distance-based metrics, as it redefines \iou\ by removing the inner bulk of the segmentation (i.e.\ hollowing), making it more sensitive to boundary variations~\cite{Cheng2021BoundaryIoU}. Computing the hollowed segmentation requires to accurately determine the \textit{interior neighborhood of the boundary} $\mbufferA$ (Fig.~\ref{fig:computation-paradigms}d), making precise boundary extraction and distance computation crucial. Similarly to \nsd, \biou\ is also influenced by distance quantization, since the choice of $\tau$ directly determines the thickness of the boundary region.
\subsection{Calculation of distances in \texttt{\meshmetrics}}
\begin{algorithm}[!tb]
    \caption{Calculation of distances in \meshmetricstt.}
    \label{alg:distance-calculation}
    \newcommand{\StateNoIndent}[1]{\Statex \hspace{-\algorithmicindent}#1}
\begin{small}
    \begin{algorithmic}[1]
        \StateNoIndent{\textbf{Input:} Segmentations \segA\ and \segB\ (2D or 3D, masks or meshes).}
        \StateNoIndent{\textbf{Output:} \parbox[t]{0.86\linewidth}{Sets of distances $\mDistSetAB$ and $\mDistSetBA$, and sets of corresponding boundary element sizes $\mBoundarySetA$ and $\mBoundarySetB$.\strut}}
        \vspace{1.25ex}
        \State \textbf{Step 1:} \textit{Boundary extraction}:
        \If{\segA\ and \segB\ are provided as masks}
            \State \parbox[t]{0.85\linewidth}{Using the \textit{surface nets} algorithm, convert \segA\ and \segB\ to\strut}
            \If{2D} 
                \State \parbox[t]{0.85\linewidth}
                {line polygons.\strut}
            \ElsIf{3D}
                \State \parbox[t]{0.85\linewidth}
                {triangle meshes.\strut}
            \EndIf
        \EndIf
        \vspace{2.25ex}
        \State \textbf{Step 2:} \textit{Calculate distances and boundary element sizes}:
        \State \parbox[t]{0.85\linewidth}
        {Initialize empty sets $\mDistSetAB$ and $\mBoundarySetA$.\strut}
        \For{each boundary element $\delta a_i \in \partial A$}
            \State \parbox[t]{0.85\linewidth}
            {Define query point $q_i$ as}
            \If{2D} 
                \State \parbox[t]{0.85\linewidth}{the midpoint of line segment $\delta a_i$.\strut} 
            \ElsIf{3D}
                \State \parbox[t]{0.85\linewidth}{the centroid of mesh triangle $\delta a_i$.\strut}
            \EndIf
            \State \parbox[t]{0.9\linewidth}{Compute distance $d_i$ from $q_i$ to $\partial B$ and append to $\mDistSetAB$.\strut}
            \State \parbox[t]{0.9\linewidth}{Compute boundary element size of $\delta a_i$ and append to $\mBoundarySetA$.\strut}
        \EndFor
        \State Repeat \textbf{Step 2} for each $\delta b_i \in \partial B$ to obtain $\mDistSetBA$ and $\mBoundarySetB$.
        \vspace{2.25ex}
        \State \textbf{Step 3:} \textit{Sort $\{\mDistSetAB,\mBoundarySetA\}$}:
        \State Sort the distances in $\mDistSetAB$ in ascending order.
        \State Apply the same order to boundary element sizes in $\mBoundarySetA$.
        \State Repeat \textbf{Step 3} to sort $\{\mDistSetBA,\mBoundarySetB\}$.
        \vspace{2.25ex}
        \StateNoIndent{\textbf{Return:} \textit{$\{\mDistSetAB,\mBoundarySetA\}$ and $\{\mDistSetBA,\mBoundarySetB\}$ for metrics computation.}}
    \end{algorithmic}
\end{small}
\end{algorithm}
The comparison of computational paradigms in Fig.~\ref{fig:computation-paradigms} highlights that mesh-based metric calculation provides a mathematically consistent discretization of the continuous formulation: as boundary elements become smaller, the discrete result converges to the continuous solution. In contrast, grid-based approaches generally lack such convergence, since they represent the boundary by non-uniformly distributed points (e.g.\ near corners), which introduces discretization artifacts. This theoretical consistency, combined with improved numerical precision, motivates distance-based metric calculations in the mesh domain, as detailed in the remainder of this section. The core process of distance calculation employed in \meshmetricstt\ is outlined in Algorithm~\ref{alg:distance-calculation}. Given a segmentation in the mask format, the first step applies the \textit{surface nets} isosurfacing algorithm (i.e.\ meshing)~\cite{Gibson1998_SurfaceNets} to obtain a mesh representation. For each mesh, we traverse its boundary elements (line segments/surface triangles in 2D/3D), compute the centroid of each element serving as the query point, and then determine the distance $\mdistAB$ from the query point to the opposing mesh. Additionally, we compute the corresponding boundary element size $\delta a$ (line segment length/surface triangle area in 2D/3D). The procedure is repeated for all boundary elements in both meshes, yielding two sets of distances $\mDistSetAB$ and $\mDistSetBA$, containing directed distances $\mdistAB$ and $\mdistBA$, respectively, and corresponding two sets $\mBoundarySetA$ and $\mBoundarySetB$, containing boundary element sizes $\delta a$ and $\delta b$, respectively. Each set of distances is sorted in ascending order, with the same ordering applied to the set of corresponding boundary element sizes. Together, these four sets form the basis for calculating the distance-based metrics (Fig.~\ref{fig:computation-paradigms}e)\footnote{For the full implementation details of all metrics, please refer to our GitHub repository, script \href{https://github.com/gasperpodobnik/MeshMetrics/blob/main/MeshMetrics/metrics.py}{\texttt{MeshMetrics/metrics.py}.}}.
\subsection{Experimental design}
To quantify the impact of different computational paradigms, we conducted an empirical analysis on 2D and 3D segmentation use cases using publicly available datasets. For the 2D analysis, we selected the \textit{INbreast dataset}\footnote{\hrefx{https://www.kaggle.com/datasets/tommyngx/inbreast2012}}~\cite{Moreira2012INbreast}, which includes 115 high-resolution mammograms with corresponding manual tumor segmentations. An additional set of segmentations was generated using an in-house method, resulting in $80$ pairs of non-empty segmentation masks~\cite{Podobnik2024-UnderstandingPitfalls-arxiv}. For the 3D analysis, we used a subset of the \textit{HaN-Seg dataset}\footnote{\hrefx{https://doi.org/10.5281/zenodo.7442914}}~\cite{Podobnik2023-Han-Seg-Dataset}, comprising $30$ CT and MR image pairs from the radiotherapy planning workflow. Each image was paired with segmentations of up to $30$ organs-at-risk, independently delineated by two clinical experts, yielding 1,559 pairs of non-empty segmentation masks~\cite{Podobnik2024-UnderstandingPitfalls-arxiv}. 
\par
We considered three sets of different element sizes for each dataset: two isotropic (a \textit{vanilla scenario} with unit size and a non-unit size) and one anisotropic. For the 2D analysis, we used \spacingOneOne, \spacingPointZeroSevenPointZeroSeven, representing a common isotropic pixel size in mammography, and ~\spacingPointZeroSevenOne. For the 3D analysis, we used \spacingOneOneOne, \spacingTwoTwoTwo, and \spacingHalfHalfTwo, representing voxel sizes typical of radiotherapy workflows and reflecting evaluation settings commonly encountered in practice. The empirical analysis involved loading the original binary segmentation masks at their native resolution, resampling them to each of the three pixel/voxel sizes, and computing distance-based metrics (\hd, \hdpercp, \masd, \assd, \nsd, \biou).  
\par
For metrics requiring user-defined parameters, i.e.\ the percentile $p$ for \hdpercp, and the distance tolerance $\tau$ for \nsd\ and \biou, we performed experiments with commonly used values of \hdpercn{\!$p$\,{=}\,95}, \nsd$_{\tau\,{=}\,2\,\textnormal{mm}}$, and \biou$_{\tau\,{=}\,2\,\textnormal{mm}}$~\cite{Maier-Hein2024-Metrics-Reloaded}. To validate \meshmetricstt, we compared it with two open-source implementations, \googledeepmindtt\ and \metricsreloadedtt, which were selected according to our implementation pitfall study~\cite{Podobnik2024-UnderstandingPitfalls-arxiv}. \googledeepmindtt\ was identified as the most sophisticated grid-based implementation due to its hybrid grid/mesh calculation principle. It uses the grid as the main representation but approximates boundary element sizes using the \textit{marching cubes} algorithm, enabling a form of pseudo-boundary weighting\footnote{Because the boundary element sizes do not exactly match but only approximate the corresponding boundary size~\cite{Podobnik2024-UnderstandingPitfalls-arxiv}.}~\cite{Nikolov2021-DeepMind}. On the other hand, \metricsreloadedtt\ provides the most complete set of metrics and serves as the official implementation of the established \metricsreloadedit\ consortium~\cite{Maier-Hein2024-Metrics-Reloaded}. To enable a consistent comparison, we implemented \assd\ and \biou\ that are not originally included in \googledeepmindtt\ using its available Python functions. Moreover, we fixed the bug in \metricsreloadedtt\ that ignored the pixel/voxel size for \biou\ calculation~\cite{Podobnik2024-UnderstandingPitfalls-arxiv} to focus on the conceptual differences in implementations rather than their coding errors. For all three implementations, we used their latest versions available as of September 2025\footnote{\texttt{\footnotesize\googledeepmind} (GitHub: \hrefx{https://github.com/google-deepmind/surface-distance}, v0.1, commit hash: \texttt{1f805ce}); \texttt{\footnotesize\metricsreloaded} (GitHub: \hrefx{https://github.com/Project-MONAI/MetricsReloaded}, v0.1.0, commit hash: \texttt{cb38dfc}); \texttt{\footnotesize\meshmetrics} (GitHub: \hrefx{https://github.com/gasperpodobnik/MeshMetrics}, v0.2.0, commit hash: \texttt{f530042}).}.
\begin{table}[!t]
    \caption{The deviations in the distance-based metrics for the 2D and 3D datasets, comparing \googledeepmindtt\ and \metricsreloadedtt\ against the proposed \meshmetricstt\ implementation, reported for two isotropic and one anisotropic voxel size as the range $\min{|}\max$ and mean $\pm$ standard deviation (SD). Deviations for \hd, \hdpercn{95}, \masd, and \assd\ are reported in millimeters (mm), while deviations for \nsd\ and \biou\ are expressed in percentage points (\%pt).}
    \label{tab:empirical-results}
\vspace{-2pt}
\centering
\renewcommand{\arraystretch}{1.05}
\setlength{\tabcolsep}{0pt}
\newcommand{\colS}{\hspace*{0.1cm}}
\newcommand{\colT}{1.5cm}
\newcolumntype{x}[1]{>{\centering\arraybackslash\hspace{0pt}}p{#1}}
\newcolumntype{L}[1]{>{\raggedright\let\newline\\\arraybackslash\hspace{0pt}}m{#1}}
\newcolumntype{C}[1]{>{\centering\let\newline\\\arraybackslash\hspace{0pt}}m{#1}}
\def\clen{2.0cm}
\newcommand{\sm}{\scalebox{0.5}[1.0]{\( - \)}}
\newcommand{\fm}{~\,}
\renewcommand{\bf}[1]{\mathbf{#1}}

\def\minmax{~~Min\,$|$\,Max}
\def\meanstd{Mean$\,{\pm}\,$SD}
\begin{small}
\begin{tabularx}{\linewidth}{L{1.7cm} l r C{\clen}C{\clen} r C{\clen}C{\clen} r C{\clen}C{\clen}}
\hline \noalign{\smallskip}
Metric & Open-source tool & \colS & \multicolumn{1}{c}{\minmax} & \multicolumn{1}{c}{\meanstd} & \colS & \minmax & \meanstd & \colS & \minmax & \meanstd \\
\noalign{\smallskip} \cline{4-5} \cline{7-8} \cline{10-11} \noalign{\bigskip}
\multicolumn{2}{l}{\textbf{2D dataset:} INbreast, pixel size:} & & 
\multicolumn{2}{c}{{(1.0, 1.0) mm}} & &
\multicolumn{2}{c}{{(0.07, 0.07) mm}} & &
\multicolumn{2}{c}{{(0.07, 1.0) mm}} \\
\noalign{\smallskip} \hline \noalign{\smallskip}
\multirow{2}{*}{$\bf{HD}$} & \googledeepmindtt       &  & $\bf{0.00}\,{|}\,\bf{0.36}$ & $\bf{0.16}\,{\pm}\,\bf{0.10}$ &  & $\bf{0.00}\,{|}\,\bf{0.02}$ & $\bf{0.01}\,{\pm}\,\bf{0.01}$ &  & $\bf{0.00}\,{|}\,\bf{0.03}$ & $\bf{0.01}\,{\pm}\,\bf{0.01}$ \\	
& \metricsreloadedtt      &  & $\bf{0.00}\,{|}\,\bf{0.36}$ & $\bf{0.16}\,{\pm}\,\bf{0.10}$ &  & $\bf{0.00}\,{|}\,\bf{0.02}$ & $\bf{0.01}\,{\pm}\,\bf{0.01}$ &  & $\bf{0.00}\,{|}\,\bf{0.03}$ & $\bf{0.01}\,{\pm}\,\bf{0.01}$ \\	
\noalign{\smallskip} \hline \noalign{\smallskip}
\multirow{2}{*}{$\bf{HD_{95}}$} & \googledeepmindtt       &  & ${\sm}\bf{6.73}\,{|}\,\bf{0.65}$ & ${\sm}\bf{0.08}\,{\pm}\,\bf{0.84}$ &  & ${\sm}\bf{0.24}\,{|}\,\bf{0.56}$ & $\bf{0.03}\,{\pm}\,\bf{0.13}$ &  & ${\sm}\bf{0.49}\,{|}\,\bf{0.48}$ & $\bf{0.02}\,{\pm}\,\bf{0.21}$ \\	
& \metricsreloadedtt      &  & ${\sm}84.8\,{|}\,2.15$ & ${\sm}2.92\,{\pm}\,12.0$ &  & ${\sm}1.68\,{|}\,1.13$ & $0.04\,{\pm}\,0.32$ &  & ${\sm}75.6\,{|}\,1.60$ & ${\sm}3.32\,{\pm}\,11.8$ \\	
\noalign{\smallskip} \hline \noalign{\smallskip}
\multirow{2}{*}{$\bf{MASD}$} & \googledeepmindtt       &  & ${\sm}\bf{0.90}\,{|}\,\bf{1.83}$ & ${\sm}\bf{0.03}\,{\pm}\,\bf{0.41}$ &  & ${\sm}\bf{0.41}\,{|}\,\bf{2.47}$ & $\bf{0.14}\,{\pm}\,\bf{0.47}$ &  & ${\sm}\bf{0.17}\,{|}\,\bf{0.25}$ & ${\sm}\bf{0.03}\,{\pm}\,\bf{0.06}$ \\	
& \metricsreloadedtt      &  & ${\sm}6.89\,{|}\,5.61$ & ${\sm}0.80\,{\pm}\,1.53$ &  & ${\sm}0.83\,{|}\,3.83$ & $0.20\,{\pm}\,0.80$ &  & ${\sm}8.09\,{|}\,5.09$ & ${\sm}1.15\,{\pm}\,2.08$ \\	
\noalign{\smallskip} \hline \noalign{\smallskip}
\multirow{2}{*}{$\bf{ASSD}$} & \googledeepmindtt       &  & ${\sm}\bf{1.74}\,{|}\,\bf{4.40}$ & ${\sm}\bf{0.03}\,{\pm}\,\bf{0.85}$ &  & ${\sm}\bf{0.81}\,{|}\,\bf{5.77}$ & $\bf{0.24}\,{\pm}\,\bf{0.98}$ &  & ${\sm}\bf{0.31}\,{|}\,\bf{0.57}$ & ${\sm}\bf{0.03}\,{\pm}\,\bf{0.12}$ \\	
& \metricsreloadedtt      &  & ${\sm}11.1\,{|}\,12.0$ & ${\sm}1.37\,{\pm}\,2.63$ &  & ${\sm}1.69\,{|}\,8.70$ & $0.34\,{\pm}\,1.66$ &  & ${\sm}15.0\,{|}\,11.0$ & ${\sm}2.12\,{\pm}\,3.80$ \\	
\noalign{\smallskip} \hline \noalign{\smallskip}
\multirow{2}{*}{$\bf{NSD_{2}}$} & \googledeepmindtt       &  & ${\sm}\bf{4.01}\,{|}\,\bf{2.46}$ & $\bf{0.42}\,{\pm}\,\bf{1.05}$ &  & ${\sm}\bf{3.30}\,{|}\,\bf{1.06}$ & ${\sm}\bf{0.21}\,{\pm}\,\bf{0.70}$ &  & ${\sm}\bf{1.45}\,{|}\,\bf{0.72}$ & ${\sm}\bf{0.14}\,{\pm}\,\bf{0.38}$ \\	
& \metricsreloadedtt      &  & ${\sm}8.82\,{|}\,9.97$ & $1.52\,{\pm}\,3.11$ &  & ${\sm}5.43\,{|}\,2.28$ & ${\sm}0.26\,{\pm}\,1.24$ &  & ${\sm}4.72\,{|}\,15.6$ & $2.68\,{\pm}\,4.02$ \\	
\noalign{\smallskip} \hline \noalign{\smallskip}
\multirow{2}{*}{$\bf{BIoU_{2}}$} & \googledeepmindtt       &  & ${\sm}6.09\,{|}\,5.78$ & ${\sm}1.91\,{\pm}\,2.27$ &  & ${\sm}0.25\,{|}\,0.39$ & ${\sm}\bf{0.06}\,{\pm}\,\bf{0.11}$ &  & ${\sm}4.06\,{|}\,\bf{2.19}$ & ${\sm}1.19\,{\pm}\,1.37$ \\	
& \metricsreloadedtt      &  & $\bf{0.00}\,{|}\,\bf{0.00}$ & $\bf{0.00}\,{\pm}\,\bf{0.00}$ &  & ${\sm}\bf{0.01}\,{|}\,\bf{0.23}$ & $0.09\,{\pm}\,0.05$ &  & $\bf{0.00}\,{|}\,2.28$ & $\bf{0.93}\,{\pm}\,\bf{0.50}$ \\	
\noalign{\smallskip} \hline \noalign{\bigskip}
\multicolumn{2}{l}{\textbf{3D dataset:} HaN-Seg, voxel size:} & & 
\multicolumn{2}{c}{{(1.0, 1.0, 1.0) mm}} & &
\multicolumn{2}{c}{{(2.0, 2.0, 2.0) mm}} & &
\multicolumn{2}{c}{{(0.5, 0.5, 2.0) mm}} \\
\noalign{\smallskip} \hline \noalign{\smallskip}
\multirow{2}{*}{$\bf{HD}$} & \googledeepmindtt       &  & ${\sm}\bf{0.17}\,{|}\,\bf{0.46}$ & $\bf{0.19}\,{\pm}\,\bf{0.12}$ &  & ${\sm}\bf{0.61}\,{|}\,\bf{0.94}$ & $\bf{0.35}\,{\pm}\,\bf{0.24}$ &  & ${\sm}0.61\,{|}\,\bf{0.36}$ & $\bf{0.12}\,{\pm}\,\bf{0.09}$ \\	
& \metricsreloadedtt      &  & ${\sm}0.45\,{|}\,0.54$ & $0.20\,{\pm}\,0.12$ &  & ${\sm}0.82\,{|}\,1.45$ & $0.38\,{\pm}\,0.25$ &  & ${\sm}\bf{0.39}\,{|}\,1.23$ & $0.14\,{\pm}\,0.10$ \\	
\noalign{\smallskip} \hline \noalign{\smallskip}
\multirow{2}{*}{$\bf{HD_{95}}$} & \googledeepmindtt       &  & ${\sm}\bf{0.67}\,{|}\,\bf{0.70}$ & $\bf{0.02}\,{\pm}\,\bf{0.18}$ &  & ${\sm}\bf{2.88}\,{|}\,\bf{1.33}$ & ${\sm}\bf{0.02}\,{\pm}\,\bf{0.41}$ &  & ${\sm}\bf{0.79}\,{|}\,\bf{1.33}$ & $\bf{0.08}\,{\pm}\,\bf{0.24}$ \\	
& \metricsreloadedtt      &  & ${\sm}4.83\,{|}\,2.58$ & $0.23\,{\pm}\,0.31$ &  & ${\sm}7.68\,{|}\,2.84$ & $0.34\,{\pm}\,0.57$ &  & ${\sm}5.31\,{|}\,8.63$ & $0.49\,{\pm}\,0.88$ \\	
\noalign{\smallskip} \hline \noalign{\smallskip}
\multirow{2}{*}{$\bf{MASD}$} & \googledeepmindtt       &  & ${\sm}\bf{0.27}\,{|}\,\bf{0.59}$ & ${\sm}\bf{0.02}\,{\pm}\,\bf{0.04}$ &  & ${\sm}\bf{0.65}\,{|}\,\bf{0.94}$ & ${\sm}\bf{0.12}\,{\pm}\,\bf{0.08}$ &  & ${\sm}\bf{0.22}\,{|}\,\bf{0.29}$ & ${\sm}\bf{0.07}\,{\pm}\,\bf{0.04}$ \\	
& \metricsreloadedtt      &  & ${\sm}0.35\,{|}\,1.55$ & $0.19\,{\pm}\,0.12$ &  & ${\sm}1.62\,{|}\,2.75$ & $0.29\,{\pm}\,0.23$ &  & ${\sm}2.12\,{|}\,1.92$ & $0.17\,{\pm}\,0.30$ \\	
\noalign{\smallskip} \hline \noalign{\smallskip}
\multirow{2}{*}{$\bf{ASSD}$} & \googledeepmindtt       &  & ${\sm}\bf{0.35}\,{|}\,\bf{1.18}$ & ${\sm}\bf{0.02}\,{\pm}\,\bf{0.06}$ &  & ${\sm}\bf{0.84}\,{|}\,\bf{1.77}$ & ${\sm}\bf{0.12}\,{\pm}\,\bf{0.09}$ &  & ${\sm}\bf{0.36}\,{|}\,\bf{0.60}$ & ${\sm}\bf{0.07}\,{\pm}\,\bf{0.05}$ \\	
& \metricsreloadedtt      &  & ${\sm}0.63\,{|}\,2.84$ & $0.20\,{\pm}\,0.15$ &  & ${\sm}2.45\,{|}\,4.86$ & $0.31\,{\pm}\,0.28$ &  & ${\sm}4.07\,{|}\,3.02$ & $0.19\,{\pm}\,0.36$ \\	
\noalign{\smallskip} \hline \noalign{\smallskip}
\multirow{2}{*}{$\bf{NSD_{2}}$} & \googledeepmindtt       &  & ${\sm}\bf{2.31}\,{|}\,8.98$ & $\bf{0.72}\,{\pm}\,\bf{1.00}$ &  & ${\sm}\bf{0.43}\,{|}\,18.6$ & $3.58\,{\pm}\,2.79$ &  & ${\sm}\bf{6.69}\,{|}\,\bf{8.23}$ & $\bf{0.46}\,{\pm}\,\bf{1.08}$ \\	
& \metricsreloadedtt      &  & ${\sm}14.4\,{|}\,\bf{6.07}$ & ${\sm}1.99\,{\pm}\,2.01$ &  & ${\sm}24.4\,{|}\,\bf{8.19}$ & ${\sm}\bf{1.04}\,{\pm}\,\bf{2.93}$ &  & ${\sm}19.1\,{|}\,9.51$ & ${\sm}3.06\,{\pm}\,3.90$ \\	
\noalign{\smallskip} \hline \noalign{\smallskip}
\multirow{2}{*}{$\bf{BIoU_{2}}$} & \googledeepmindtt       &  & ${\sm}8.24\,{|}\,4.49$ & ${\sm}1.12\,{\pm}\,1.42$ &  & ${\sm}24.8\,{|}\,12.1$ & ${\sm}\bf{9.21}\,{\pm}\,\bf{5.88}$ &  & ${\sm}10.1\,{|}\,7.04$ & ${\sm}1.29\,{\pm}\,2.11$ \\	
& \metricsreloadedtt      &  & $\bf{0.00}\,{|}\,\bf{0.00}$ & $\bf{0.00}\,{\pm}\,\bf{0.00}$ &  & ${\sm}\bf{23.0}\,{|}\,\bf{2.44}$ & ${\sm}9.91\,{\pm}\,5.80$ &  & $\bf{0.00}\,{|}\,\bf{0.00}$ & $\bf{0.00}\,{\pm}\,\bf{0.00}$ \\	
\noalign{\smallskip} \hline
\end{tabularx}
\end{small}
    \vspace{-3ex}
\end{table}
\section{Results}
\subsection{Metric value variability}
\begin{figure*}[!b]
    \centering
    \includegraphics[width=\textwidth]{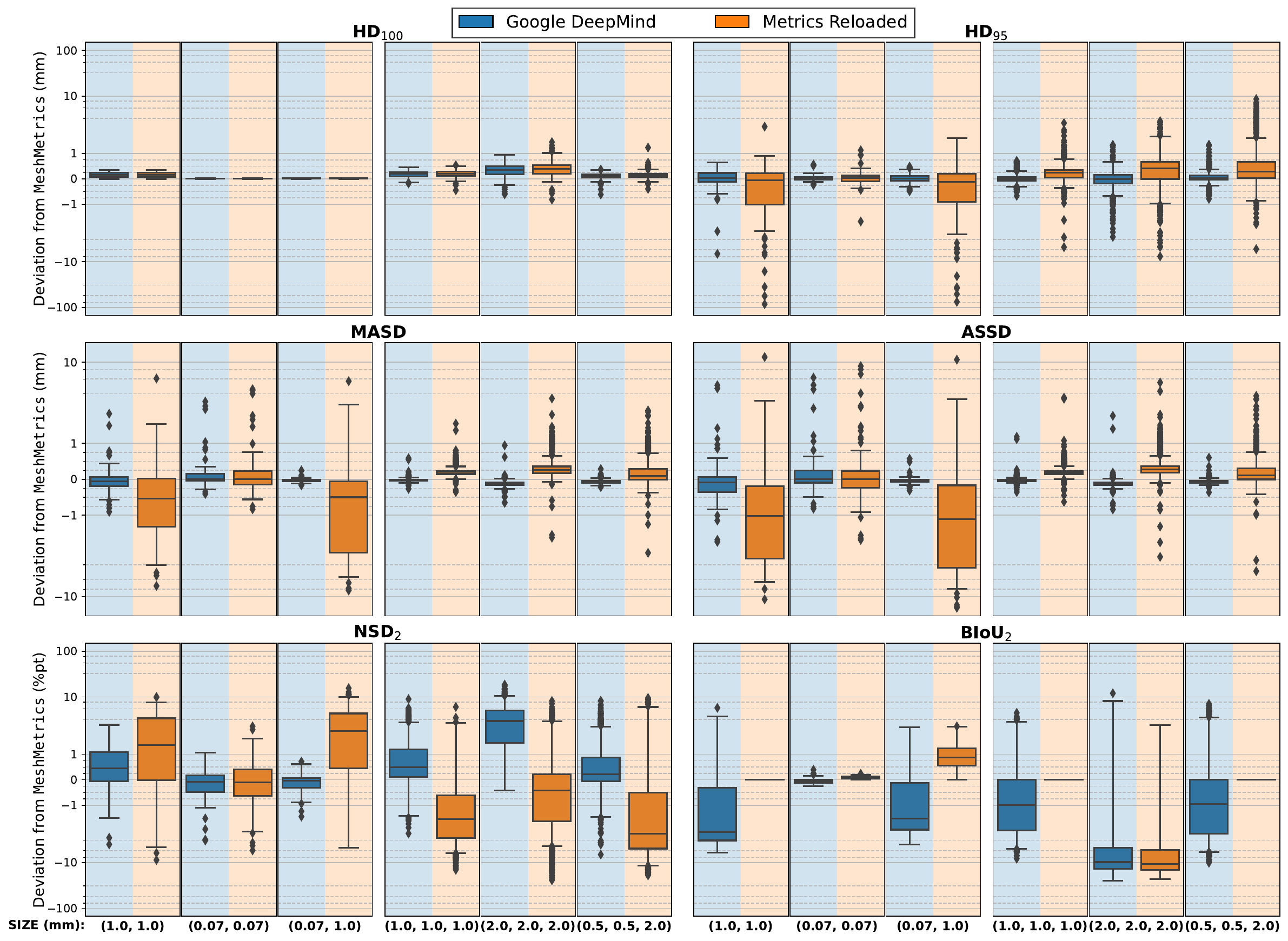}
    \caption{The deviations in the distance-based metrics on the 2D and 3D datasets, as obtained by \googledeepmindtt\ and \metricsreloadedtt\ against the reference \meshmetricstt\, reported for two isotropic and one anisotropic pixel/voxel size in the form of box-plots. Note that a linear scale is applied to values within the interval $[-1,1]$, and a symmetric logarithmic scale is used for values outside this interval to better visualize both smaller and larger differences.}
    \label{fig:empirical-boxplots}
\end{figure*}

As a direct comparison of raw metric values across each dataset would not be meaningful, we report instead the deviations of \googledeepmindtt\ and \metricsreloadedtt\ against \meshmetricstt\ serving as a reference: $m_{j,i} - m_{\textnormal{MM},i}$, where $m_{j,i}$ is the metric value from the $j$-th open-source tool (\googledeepmindtt\ or \metricsreloadedtt), and $m_{\textnormal{MM},i}$ is the reference value from \meshmetricstt\ for the $i$-th pair of segmentations. For \textit{absolute metrics} measured in metric units (\hd, \hdpercp, \masd, \assd), deviations are reported in millimeters (mm), while \textit{relative metrics} (\nsd, \biou) are reported in percentage points (\%pt). When observed against \meshmetricstt, positive deviations indicate an \textit{overestimation} of the metric (i.e.\ over-pessimistic for absolute and over-optimistic for relative metrics), whereas negative deviations indicate an \textit{underestimation} (i.e.\ over-optimistic for absolute and over-pessimistic for relative metrics). Although both types are undesirable, over-optimistic estimates are particularly concerning, as they can lead to misleading conclusions, for example, when compared to values reported in the literature. The quantitative results for the 2D and 3D analysis, stratified by pixel/voxel size, are summarized in Table~\ref{tab:empirical-results} and Fig.~\ref{fig:empirical-boxplots}, and revealed notable differences across all metrics, except for \hd\ (i.e.\ \hd$_{100}$), where all deviations are relatively small.
\subsection{Edge case handling}
An important but often overlooked aspect of metric calculation is the consistent handling of edge cases, i.e.\ situations where one or both input segmentations are empty. We observed discrepancies not only between \googledeepmindtt\ and \metricsreloadedtt, but also among different metrics within the same tool. These findings, together with the proposed behavior of our implementation, are summarized in Table~\ref{tab:edge-cases}. Specifically, when exactly one segmentation is empty, \meshmetricstt\ returns the worst possible value for the given metric (i.e.\ $\infty$ for absolute and $0$\% for relative metrics), whereas when both segmentations are empty, it returns the best possible value (i.e.\ $0$\,mm for absolute and $100$\% for relative metrics). To further support robust result aggregation, \meshmetricstt\ raises warnings to inform the user about the edge case scenario and additionally provides flags that indicate which segmentation is empty, thereby enabling downstream performance analyses (i.e.\ false positives, false negatives, and true negatives).

\begin{table*}[!t]
    \caption{Edge case handling analysis: XOR denotes the case where exactly one of the two segmentations is empty, while AND denotes the case where both are empty. In addition to the distance-based metrics, \dsc\ is included to provide a more comprehensive overview.}
    \label{tab:edge-cases}

\centering
\newcommand{\toolNameDOIX}[2]{\href{#2}{\texttt{#1}}}
\def\nameDeepMind{\toolNameDOIX{Google DeepMind}{https://github.com/google-deepmind/surface-distance}}
\def\nameMetricsReloaded{\toolNameDOIX{Metrics Reloaded}{https://github.com/Project-MONAI/MetricsReloaded}}
\def\nameMeshMetrics{\toolNameDOIX{MeshMetrics}{https://github.com/gasperpodobnik/MeshMetrics}}

\newcommand{\cellfmt}[2]{\footnotesize{#1} & \footnotesize{#2}}

\newcolumntype{C}[1]{>{\centering\let\newline\\\arraybackslash\hspace{0pt}}m{#1}}
\newcolumntype{Y}{>{\centering\arraybackslash}X}
\renewcommand{\arraystretch}{1.2} 
\small
\def\Off{\textsuperscript{~\,}} 
\newcommand{\sm}{\scalebox{0.75}[1.0]{\( - \)}}
\def\Nan{\texttt{NaN}} 
\def\Inf{\texttt{$\infty$}} 
\def\Warn{\textsuperscript{\texttt{W}}} 
\setlength{\tabcolsep}{0pt}
\def\colM{2.0cm}
\def\colB{0.1cm}
%
\begin{tabularx}{\linewidth}{
l C{\colB}  
*{2}{Y} C{\colB} 
*{2}{Y} C{\colB} 
*{2}{Y} C{\colB} 
*{2}{Y} C{\colB} 
*{2}{Y} C{\colB} 
*{2}{Y} C{\colB} 
}
\hline \noalign{\smallskip}
&&
\multicolumn{2}{C{\colM}}{\dsc\ (\%)}     &&
\multicolumn{2}{C{\colM}}{\hdpercp\ (mm)} &&
\multicolumn{2}{C{\colM}}{\masd\ (mm)}    &&
\multicolumn{2}{C{\colM}}{\assd\ (mm)}    &&
\multicolumn{2}{C{\colM}}{\nsd\ (\%)}	  && 
\multicolumn{2}{C{\colM}}{\biou\ (\%)}    \\
\cline{3-4} \cline{6-7} \cline{9-10} \cline{12-13} \cline{15-16} \cline{18-19} \noalign{\smallskip}
Open-source tool   &&
\cellfmt{XOR}{AND} &&
\cellfmt{XOR}{AND} &&
\cellfmt{XOR}{AND} &&
\cellfmt{XOR}{AND} &&
\cellfmt{XOR}{AND} &&
\cellfmt{XOR}{AND} \\
\noalign{\smallskip} \hline \noalign{\smallskip}
%
\googledeepmindtt\ &&
\cellfmt{0}{\Nan}		   &&
\cellfmt{\Inf}{\Inf}	   &&
\cellfmt{\Nan}{\Nan}	   &&
\cellfmt{\ddag}{\ddag}	   &&
\cellfmt{0}{\Off\Nan\Warn} &&
\cellfmt{\ddag}{\ddag}	   \\
%
\metricsreloadedtt\ &&
\cellfmt{0}{\Off100\Warn}	            &&
\cellfmt{\Off\Nan\Warn}{\Off0\Warn}	&&
\cellfmt{\Off\Nan\Warn}{\Off0\Warn}	&&
\cellfmt{\Off\Nan\Warn}{\Off0\Warn}	&&
\cellfmt{0}{\Off100\Warn}             &&
\cellfmt{0}{\Off100\Warn} \\
\noalign{\smallskip} \hline \noalign{\smallskip}
%
\meshmetricstt\ &&
\cellfmt{\Off0\Warn}{\Off100\Warn}    &&
\cellfmt{\Off\Inf\Warn}{\Off0\Warn} &&
\cellfmt{\Off\Inf\Warn}{\Off0\Warn} &&
\cellfmt{\Off\Inf\Warn}{\Off0\Warn} &&
\cellfmt{\Off0\Warn}{\Off100\Warn}    &&
\cellfmt{\Off0\Warn}{\Off100\Warn}   \\
\noalign{\smallskip} \hline
\multicolumn{19}{p{0.985\textwidth}}{\footnotesize{%
    \Nan: not a number;
    \texttt{W}: additional warning message;
    \ddag: {\footnotesize \texttt{\googledeepmind}} does not natively support \assd\ and \biou; therefore, their analysis is omitted.

}}
\end{tabularx}
    \vspace{-3ex}
\end{table*}
\subsection{Computational efficiency}
\begin{figure}[!hb]
    \centering
    \includegraphics[width=0.58\textwidth]{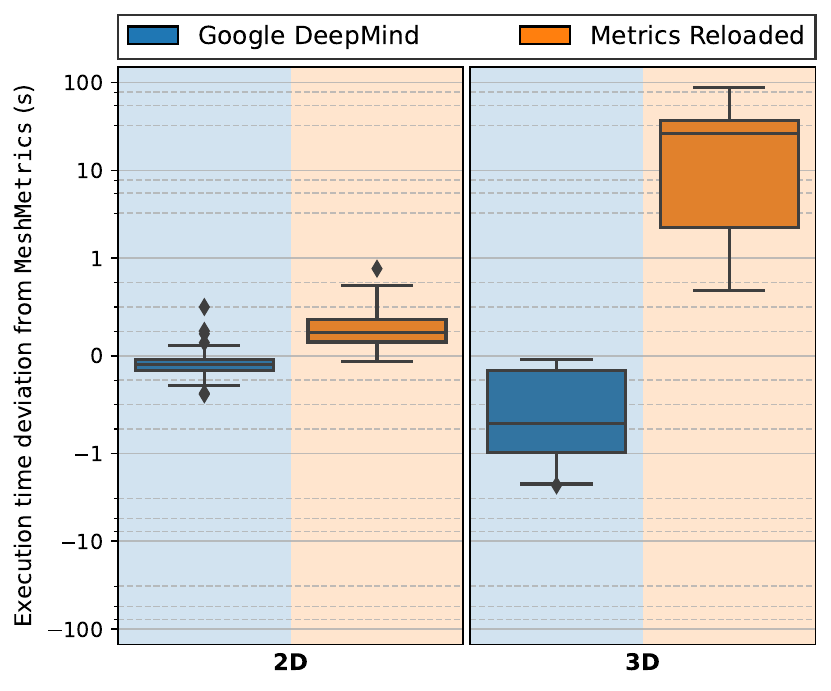}
    \caption{Comparison of the \hd\ computational efficiency on 2D and 3D datasets, reported as runtime deviations from \meshmetricstt\ for the pixel/voxel size of \spacingPointZeroSevenPointZeroSeven\ and \spacingOneOneOne. Note that a linear scale is applied to values within the interval $[-1,1]$, and a symmetric logarithmic scale is used for values outside this interval to better visualize both smaller and larger differences.}
    \label{fig:compute-efficiency}
    \vspace{-3ex}
\end{figure}

For the computational efficiency assessment, we focused on \hd, since it contains most computational steps shared by all distance-based metrics, with the main cost coming from distance calculation. Additionally, in the case of \meshmetricstt, meshing was performed internally, and its computational cost was therefore included in the efficiency analysis. In 2D, we used the original high-resolution mammograms with pixel size of \spacingPointZeroSevenPointZeroSeven\ (segmentation masks exceeding $2000\,{\times}\,3000$\,pixels), while in 3D, images were resampled to \spacingOneOneOne. Execution times are reported relative to \meshmetricstt\ (Fig.~\ref{fig:compute-efficiency}), with positive values indicating slower and negative values faster execution. In 2D, all implementations showed comparable runtimes: \googledeepmindtt\ was marginally faster (median $0.09$\,s), while \metricsreloadedtt\ was slightly slower (median $0.24$\,s) than \meshmetricstt. In 3D, \googledeepmindtt\ remained the most efficient, running $0.7$\,s faster, whereas \metricsreloadedtt\ was substantially slower, with a median runtime of $26.4$\,s above that of \meshmetricstt.
\section{Discussion}\label{sec:discussion}
\subsection{Metric value variability}
\subsubsection{\hd}
\hd\ is the only metric with a consistent mathematical definition across all open-source tools~\cite{Podobnik2024-UnderstandingPitfalls-arxiv}, as its aggregation is based on the maximum of the two directed distances, which avoids the need to account for boundary element sizes. This consistency is reflected in the low deviations observed in the empirical results, especially for the 2D dataset. For the 3D dataset, the deviations are relatively small but more pronounced than in 2D, as calculations in 3D are more sensitive to boundary extraction differences.    
\subsubsection{\hdpercn{95}}
In contrast to \hd, its percentile definition requires careful computation that incorporates boundary element sizes (Sec.~\ref{sec:Methods}). Among several mathematical definitions~\cite{Podobnik2024-UnderstandingPitfalls-arxiv}, we adopted the version that takes the maximum of both directed percentiles, and argue that this is most consistent with the original \hd\ definition\footnote{A ``Hausdorffian'' way of aggregating the two directed distances.}. Since \googledeepmindtt\ approximates boundary element sizes for percentile calculation and \metricsreloadedtt\ estimates the percentile based only on the set cardinality, it is not surprising that \googledeepmindtt\ performed most similarly to \meshmetricstt. However, its small errors in boundary size approximation~\cite{Podobnik2024-UnderstandingPitfalls-arxiv} still lead to occasional outliers. By contrast, the severe underestimation observed for \metricsreloadedtt, in one case exceeding 80\,mm, is alarming and underscores the importance of correct percentile computation\footnote{An in-depth analysis of the outlier resulting in an underestimation of ${-}84.4$\,mm is provided in Supplementary Materials (Note~\ref{sec:hd95-outlier} and Fig.~\ref{fig:hd95-outlier}).}.
\subsubsection{\masd\ and \assd}
These two metrics share the same aggregation principle, which explains their similar behavior. Although they are less sensitive than \hdpercp\ to precise weighting by boundary element sizes, we still observed substantial deviations, particularly for the anisotropic 2D case, where \metricsreloadedtt\ underestimated \assd\ by an average of $2.12$\,mm. Such systematic deviations are concerning, as improvements of even a single millimeter are often considered significant when evaluating automatic segmentation methods.
\subsubsection{\nsd}
Besides the effects of boundary extraction methods and different mathematical definitions, \nsd\ is also influenced by distance quantization. This issue arises from its definition, with tolerance $\tau$ specifying the maximum allowable deviation between the two boundaries. In the grid domain, quantization effects are particularly pronounced, as evidenced by the jagged curves of \googledeepmindtt\ and \metricsreloadedtt\ in Fig.~\ref{fig:quantized-distances}. By contrast, mesh-based calculation yields a smoother curve due to a more precise distance computation. Empirically, these differences led to substantial deviations (Fig.~\ref{fig:empirical-boxplots}), amounting to nearly 10\%pt in 2D and more than 24\%pt in 3D, with mean deviations of up to 3.58\%pt for \googledeepmindtt. Interestingly, the largest deviations were observed for \googledeepmindtt, despite being the official \nsd\ implementation~\cite{Nikolov2021-DeepMind}, highlighting the impact of quantized distances in the grid domain. Moreover, our results shed light on the often-overlooked issue of selecting parameter $\tau$. Since we computed \nsd\ with $\tau\,{=}\,2$\,mm, the effect is most evident for voxel size \spacingTwoTwoTwo, where $2$\,mm corresponds to the minimal possible non-zero distance. As shown in Fig.~\ref{fig:quantized-distances}, large jagged jumps in \nsd\ appear when $\tau$ is close to the voxel dimension, stabilizing only when $\tau$ exceeds the minimal voxel dimension by several multiples. 
\subsubsection{\biou}
Unlike other distance-based metrics, \biou\ is a hybrid between overlap- and distance-based metrics, as it computes \iou\ on hollowed segmentations to increase the sensitivity to boundary deviations. While distance computation is therefore less critical, consistent implementation remains important. Since the intersection and union operations are fastest in the grid domain, \meshmetricstt\ adopts a hybrid grid/mesh approach: the grid serves as the foundation, but distances are computed from the mesh boundary, thereby ensuring a consistent boundary representation across all metrics. The deviations against \googledeepmindtt\ and \metricsreloadedtt\ arose primarily from the differences in boundary extraction algorithms, and diminished with an increasing $\tau$, with all three implementations converging towards \iou\ when $\tau$ was sufficiently large (i.e.\ equal to or greater than the largest interior distance within the segmentation).
\subsection{Edge case handling}
When exactly one input segmentation is empty (cf.\ XOR in Table~\ref{tab:edge-cases}), the metrics should assume their worst values: infinity (\Inf) for absolute and zero ($0$\%) for relative metrics. When both segmentations are empty (cf.\ AND in Table~\ref{tab:edge-cases}), the two segmentations perfectly align, and the metrics should assume their best values: zero ($0$\,mm) for absolute and one ($100$\%) for relative metrics. However, these cases require careful interpretation, as a high number of empty inputs can saturate the aggregated metrics. To provide transparency and consistency, \meshmetricstt\ outputs a warning and sets corresponding flags, allowing the user to handle edge cases explicitly during post-analysis, for instance, reporting performance separately when either segmentation is empty (false positives and false negatives), or when both are empty (true negatives). Such handling is particularly important in pathology segmentation, where structures may be absent in a large proportion of images, and a careful, disentangled evaluation of detection and segmentation performance is crucial~\cite{Reinke2024-Metrics-Pitfalls, Maier-Hein2024-Metrics-Reloaded}.
\subsection{Computational efficiency}
While computational efficiency is secondary to metric reliability, it remains important in practice for enabling rapid prototyping and conserving computational resources. In 2D, \meshmetricstt\ performed \textit{on par}, whereas in 3D it was slightly slower than \googledeepmindtt, yet still considerably faster than \metricsreloadedtt. Given that \metricsreloadedtt\ is relatively slow but nevertheless widely used, this indicates that \meshmetricstt\ is sufficiently fast, particularly in light of its advantages in terms of reliability. \meshmetricstt\ is also lean in terms of dependencies, relying on only four Python packages (\texttt{\small vtk}, \texttt{\small numpy}, \texttt{\small SimpleITK}, \texttt{\small SimpleITKUtilities}), and can be easily installed by cloning and \texttt{\small pip} installing the repository. Furthermore, the input can be provided either as a segmentation mask (\texttt{\small SimpleITK.Image} or \texttt{\small numpy.ndarray}) or surface mesh (\texttt{\small vtk.vtkPolyData}), therefore ensuring that the same calculation principles apply to both grid- and mesh-based inputs.
\subsection{Limitations}
First, while not a limitation of our approach \textit{per se}, mesh-based calculations (similar to grid-domain calculations) require the selection of a boundary extraction method (i.e.\ an isosurfacing or meshing algorithm)~\cite{Schroeder2015-FlyingEdges, Grothausmann2016-DiscreteMarchingCubes, Nielson1991-MCAmbiguity, Gibson1998_SurfaceNets}. Considering the mathematical consistency with other metrics, mesh quality, and computational efficiency, \meshmetricstt\ relies on the \textit{surface nets} algorithm~\cite{Gibson1998_SurfaceNets}\footnote{An extended analysis of this choice is provided in Supplemental Materials, including examples of different meshing methods and their empirical performance (cf.\ Note~\ref{sec:meshing-methods}, Fig.~\ref{fig:meshing-methods} and Table~\ref{tab:meshing-methods}).}. Second, among the analyzed metrics, \biou\ stands out as a hybrid between overlap- and distance-based metrics. While it could be, in principle, implemented entirely in the mesh domain, this would require robust mesh thinning and boolean mesh operations, which are still under development~\cite{Sajovic2025TrueformBoolean}. \meshmetricstt\ circumvents these limitations by adopting a hybrid approach: it uses the mesh for precise boundary representation while relying on the grid for the calculation of intersection and union, which is naturally and elegantly defined in that domain. While it serves as a robust solution, this approach can be revisited once more robust algorithms become available. Third, while this study is concise in scope -- covering a set of metrics that share a common underlying calculation principle -- we acknowledge that additional metrics could be investigated and integrated into the repository. One example is the centerline Dice metric~\cite{Shit2021clDice}, which requires a different set of fundamentals (e.g.\ a robust skeletonization algorithm that is currently not yet available~\cite{Podobnik2025CenterlineDice}) and thus warrants separate in-depth study. By covering all distance-based metrics analyzed in \metricsreloadedit~\cite{Maier-Hein2024-Metrics-Reloaded}, we believe that our selection captures the most widely used distance-based metrics, and that its underlying computational principles provide a solid foundation for extending the framework to additional metrics in the future.
\section{Conclusion}
We proposed \meshmetricstt, a precise open-source implementation of distance-based metrics in the mesh domain. Through comparison with \googledeepmindtt\ and \metricsreloadedtt, we validated its correctness and demonstrated the limitations of grid-based approaches. \meshmetricstt\ is more robust to discretization artifacts, computationally comparable to \googledeepmindtt, and substantially faster than \metricsreloadedtt\ in 3D, making it also practical for segmentation validation. Along with the highlighted implementation pitfalls~\cite{Podobnik2024-UnderstandingPitfalls-arxiv}, we shed light on the challenges of segmentation evaluation, and pave the way towards a more precise and objective method validation, biomarker computation, and benchmarking.
%
\section*{Acknowledgments}
We would like to thank the authors of \googledeepmindtt\ and \metricsreloadedtt\ for publicly releasing their open-source implementations. We further thank the \metricsreloadedit\ consortium for advancing metrics standardization through guidelines, nomenclature, and unified mathematical definitions. Finally, we acknowledge \textit{Kitware, Inc.} (New York, USA) for providing open-source libraries, including \texttt{\small vtk} and \texttt{\small itk}, that form the basis of our implementation.
\par
This study was supported by the Slovenian Research and Innovation Agency (ARIS), project No.\ J2-4453, J2-50067, J2-60042 and P2-0232, and by the European Union Horizon project ARTILLERY, grant agreement No.\ 101080983.
\bibliography{main}
\clearpage
\section*{Appendix}
\setcounter{secnumdepth}{2}
\renewcommand{\thesubsection}{A\arabic{subsection}}
\setcounter{subsection}{0}
\titleformat{\subsection}
	{\normalfont\bfseries}
	{Note~\thesubsection.}{0.5em}{}[]
\renewcommand{\thefigure}{A\arabic{figure}}
\setcounter{figure}{0}
\renewcommand{\thetable}{A\arabic{table}}
\setcounter{table}{0}
%
\subsection{Example of \hdpercn{95} outlier}
\label{sec:hd95-outlier}
\begin{figure}[!h]
    \centering
    \includegraphics[width=0.98\textwidth]{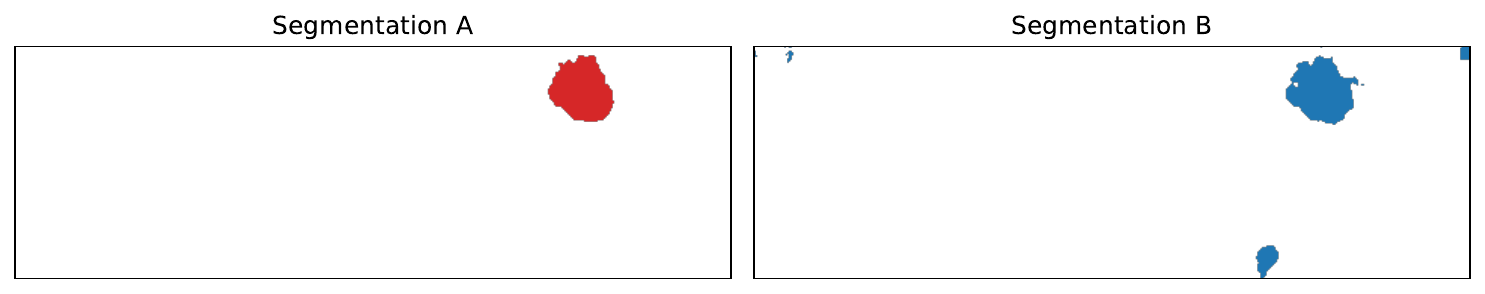}
    \caption{The 2D case resulting in a significant \hdpercn{95} underestimation of ${-}84.8$\,mm by \metricsreloadedtt\ relative to \meshmetricstt.}
    \label{fig:hd95-outlier}
    \vspace{-4ex}
\end{figure}

The example from our 2D analysis using pixel size \spacingOneOne\ that resulted in an \hdpercn{95} underestimation of ${-}84.8$\,mm by \metricsreloadedtt\  relative to \meshmetricstt, whereas \googledeepmindtt\ deviated by only ${-}0.03$\,mm (Table~\ref{tab:empirical-results}), is illustrated in Fig.~\ref{fig:hd95-outlier}. Such a large discrepancy arose because \metricsreloadedtt\ does not apply boundary weighting and thus assumes a uniform distribution of extracted points along the boundary. As one segmentation mask contains multiple connected components, and because \metricsreloadedtt\ ignores line segment lengths when computing percentiles, the 95th percentile can exhibit large jumps when its calculation is based solely on the cardinality of the set of distances. Although \hdpercn{95} is generally robust to outliers, this example shows that extreme cases can still significantly impact metric estimation. In contrast, both \meshmetricstt\ and \googledeepmindtt\ correctly captured the segmentation quality in this scenario.
\vspace{-3ex}
\subsection{Meshing methods}
\label{sec:meshing-methods}
\begin{figure}[!b]
    \centering
    \input{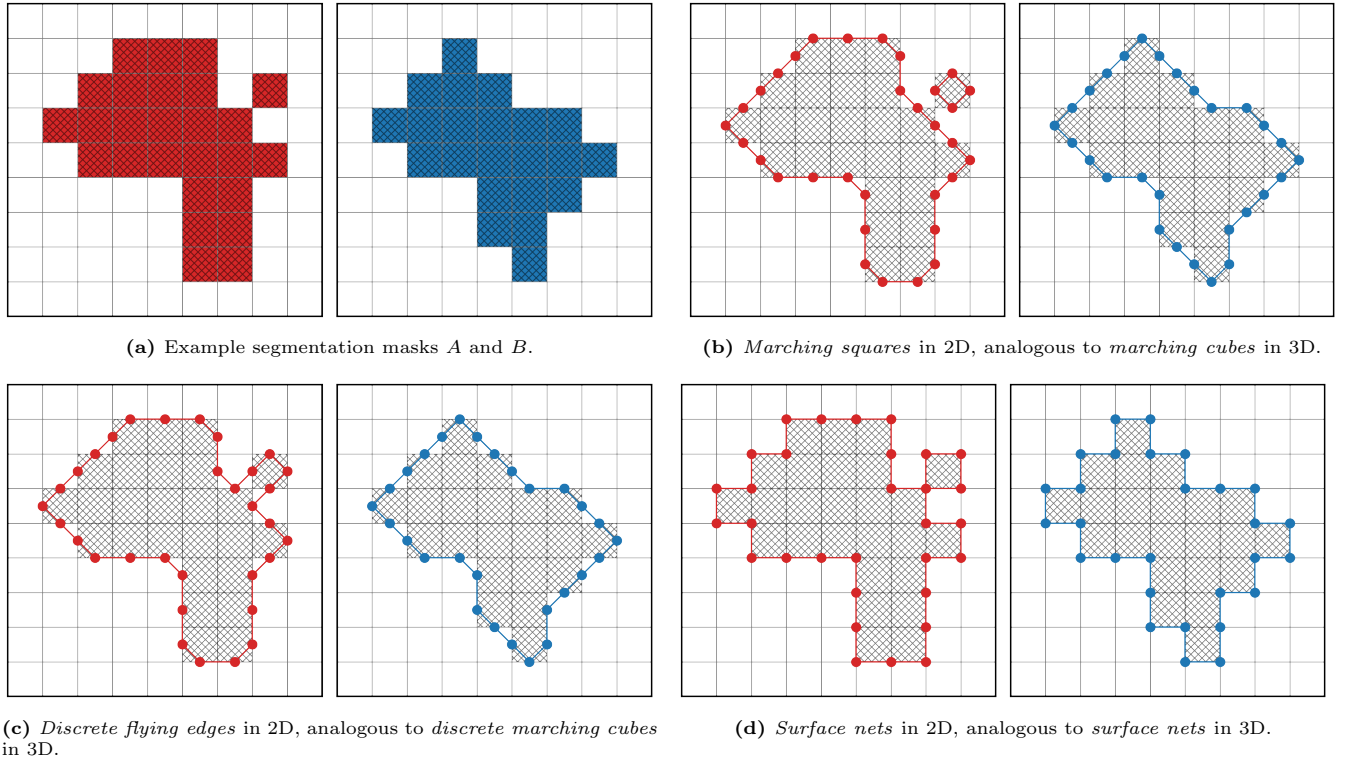}
    \caption{Comparison of different meshing methods.
    \subref{fig:meshing-methods-a}~Example segmentation masks.
    \subref{fig:meshing-methods-b}~The \textit{marching squares} / \textit{marching cubes} algorithm is the traditional and widely adopted meshing algorithm that produces smooth results.
    \subref{fig:meshing-methods-c}~The \textit{discrete flying edges} / \textit{discrete marching cubes} algorithm~\cite{Schroeder2015-FlyingEdges, Grothausmann2016-DiscreteMarchingCubes} builds upon the traditional algorithm but is non-parametric (i.e.\ does not require an isosurface value) and handles ambiguous cases more conveniently for most image segmentation tasks~\cite{Nielson1991-MCAmbiguity}, such as in the shown example, where a single instead of two connected components is produced.  
    \subref{fig:meshing-methods-d}~The \textit{surface nets} algorithm~\cite{Gibson1998_SurfaceNets} closely resembles the boundary extraction method employed by \googledeepmindtt~\cite{Podobnik2024-UnderstandingPitfalls-arxiv}. This algorithm is also used by \meshmetricstt.
    }
    \label{fig:meshing-methods}
\end{figure}

An overview of common methods for converting binary segmentation masks into meshes, often referred to as \textit{isosurface extraction} or simply \textit{meshing}, is provided in Fig.~\ref{fig:meshing-methods}. 
All methods are demonstrated using 2D binary masks with an isosurface value of $0.5$ where applicable, and their analogous 3D methods are noted. In \meshmetricstt, we use the \textit{surface nets} algorithm~\cite{Gibson1998_SurfaceNets} due to its superior speed and ability to generate boundaries that closely approximate the true segmentation in both 2D and 3D grids. This ensures a consistent coverage of the input grid, which is particularly important for the computation of specific metrics, for example, \biou\ should match IoU when $\tau$ is set to $\infty$ or a sufficiently large value. To avoid potential biases and ensure consistent results, the meshing algorithm in \meshmetricstt\ is hardcoded to \textit{surface nets} rather than being exposed as a user-specified parameter. Such a design prevents unintended modifications and maintains objectivity when using the tool.
\par
To assess the empirical impact of meshing on the resulting metric values, Table~\ref{tab:meshing-methods} presents a detailed quantitative comparison of all meshing algorithms. Similarly to Table~\ref{tab:empirical-results}, the results are reported as deviations from \meshmetricstt\ that uses the \textit{surface nets} meshing algorithm, and serves as the reference. The differences between the \textit{marching squares} and \textit{discrete flying edges} algorithms in 2D, and the \textit{marching cubes} and \textit{discrete marching cubes} algorithms in 3D are minimal, as these algorithms are essentially equivalent, differing only in how they handle ambiguous cases (cf.\ reference segmentations in Fig.~\ref{fig:meshing-methods-b}). Although systematic differences exist, the magnitude of outliers is relatively small.
\begin{table}[!h]
    \caption{The deviations in distance-based metrics obtained using the \textit{marching squares}$|$\textit{marching cubes}, and the \textit{discrete flying edges}$|$\textit{discrete marching cubes} meshing algorithms (2D$|$3D, respectively), compared to the reference \textit{surface nets} meshing algorithm employed in \meshmetricstt\ (Fig.~\ref{fig:meshing-methods}). The results are reported for two isotropic and one anisotropic voxel size as the range $\min{|}\max$ and mean $\pm$ standard deviation (SD). Deviations for \hd, \hdpercn{95}, \masd, and \assd\ are given in millimeters (mm), while deviations for \nsd\ and \biou\ are expressed in percentage points (\%pt).}
    \label{tab:meshing-methods}
\vspace{-2pt}
\centering
\renewcommand{\arraystretch}{1.05}
\setlength{\tabcolsep}{0pt}
\newcommand{\colS}{\hspace*{0.1cm}}
\newcommand{\colT}{1.5cm}
\newcolumntype{x}[1]{>{\centering\arraybackslash\hspace{0pt}}p{#1}}
\newcolumntype{L}[1]{>{\raggedright\let\newline\\\arraybackslash\hspace{0pt}}m{#1}}
\newcolumntype{C}[1]{>{\centering\let\newline\\\arraybackslash\hspace{0pt}}m{#1}}
\def\clen{2.0cm}
\newcommand{\sm}{\scalebox{0.5}[1.0]{\( - \)}}
\newcommand{\fm}{~\,}
\renewcommand{\bf}[1]{\mathbf{#1}}

\def\minmax{~~Min\,$|$\,Max}
\def\meanstd{Mean$\,{\pm}\,$SD}
\begin{small}
\begin{tabularx}{\linewidth}{L{1.7cm} l r C{\clen}C{\clen} r C{\clen}C{\clen} r C{\clen}C{\clen}}
\hline \noalign{\smallskip}
Metric & Open-source tool & \colS & \multicolumn{1}{c}{\minmax} & \multicolumn{1}{c}{\meanstd} & \colS & \minmax & \meanstd & \colS & \minmax & \meanstd \\
\noalign{\smallskip} \cline{4-5} \cline{7-8} \cline{10-11} \noalign{\bigskip}
\multicolumn{2}{l}{\textbf{2D dataset:} INbreast, pixel size:} & & 
\multicolumn{2}{c}{{(1.0, 1.0) mm}} & &
\multicolumn{2}{c}{{(0.07, 0.07) mm}} & &
\multicolumn{2}{c}{{(0.07, 1.0) mm}} \\
\noalign{\smallskip} \hline \noalign{\smallskip}
\multirow{2}{*}{$\bf{HD}$} & Marching squares &  & ${\sm}\bf{0.25}\,{|}\,\bf{0.00}$ & ${\sm}\bf{0.12}\,{\pm}\,\bf{0.07}$ &  & ${\sm}\bf{0.02}\,{|}\,\bf{0.00}$ & ${\sm}\bf{0.01}\,{\pm}\,\bf{0.00}$ &  & ${\sm}\bf{0.25}\,{|}\,\bf{0.00}$ & ${\sm}\bf{0.03}\,{\pm}\,\bf{0.05}$ \\	
& Discrete flying edges &  & ${\sm}\bf{0.25}\,{|}\,\bf{0.00}$ & ${\sm}\bf{0.12}\,{\pm}\,\bf{0.07}$ &  & ${\sm}\bf{0.02}\,{|}\,\bf{0.00}$ & ${\sm}\bf{0.01}\,{\pm}\,\bf{0.00}$ &  & ${\sm}\bf{0.25}\,{|}\,\bf{0.00}$ & ${\sm}\bf{0.03}\,{\pm}\,\bf{0.05}$ \\	
\noalign{\smallskip} \hline \noalign{\smallskip}
\multirow{2}{*}{$\bf{HD_{95}}$} & Marching squares &  & ${\sm}\bf{6.92}\,{|}\,\bf{0.65}$ & ${\sm}\bf{0.12}\,{\pm}\,\bf{0.86}$ &  & ${\sm}\bf{0.23}\,{|}\,\bf{0.56}$ & $\bf{0.03}\,{\pm}\,\bf{0.13}$ &  & ${\sm}\bf{0.25}\,{|}\,\bf{0.48}$ & $\bf{0.01}\,{\pm}\,\bf{0.15}$ \\	
& Discrete flying edges &  & ${\sm}\bf{6.92}\,{|}\,\bf{0.65}$ & ${\sm}\bf{0.12}\,{\pm}\,\bf{0.86}$ &  & ${\sm}\bf{0.23}\,{|}\,\bf{0.56}$ & $\bf{0.03}\,{\pm}\,\bf{0.13}$ &  & ${\sm}\bf{0.25}\,{|}\,\bf{0.48}$ & $\bf{0.01}\,{\pm}\,\bf{0.15}$ \\	
\noalign{\smallskip} \hline \noalign{\smallskip}
\multirow{2}{*}{$\bf{MASD}$} & Marching squares &  & ${\sm}\bf{0.89}\,{|}\,\bf{1.88}$ & ${\sm}\bf{0.00}\,{\pm}\,\bf{0.41}$ &  & ${\sm}\bf{0.41}\,{|}\,\bf{2.47}$ & $\bf{0.14}\,{\pm}\,\bf{0.47}$ &  & ${\sm}\bf{0.16}\,{|}\,\bf{0.27}$ & ${\sm}\bf{0.01}\,{\pm}\,\bf{0.06}$ \\	
& Discrete flying edges &  & ${\sm}\bf{0.89}\,{|}\,\bf{1.88}$ & ${\sm}\bf{0.00}\,{\pm}\,\bf{0.41}$ &  & ${\sm}\bf{0.41}\,{|}\,\bf{2.47}$ & $\bf{0.14}\,{\pm}\,\bf{0.47}$ &  & ${\sm}\bf{0.16}\,{|}\,\bf{0.27}$ & ${\sm}\bf{0.01}\,{\pm}\,\bf{0.06}$ \\	
\noalign{\smallskip} \hline \noalign{\smallskip}
\multirow{2}{*}{$\bf{ASSD}$} & Marching squares &  & ${\sm}\bf{1.72}\,{|}\,\bf{4.43}$ & ${\sm}\bf{0.01}\,{\pm}\,\bf{0.86}$ &  & ${\sm}\bf{0.81}\,{|}\,\bf{5.77}$ & $\bf{0.24}\,{\pm}\,\bf{0.98}$ &  & ${\sm}\bf{0.31}\,{|}\,\bf{0.59}$ & ${\sm}\bf{0.02}\,{\pm}\,\bf{0.12}$ \\	
& Discrete flying edges &  & ${\sm}\bf{1.72}\,{|}\,\bf{4.43}$ & ${\sm}\bf{0.01}\,{\pm}\,\bf{0.86}$ &  & ${\sm}\bf{0.81}\,{|}\,\bf{5.77}$ & $\bf{0.24}\,{\pm}\,\bf{0.98}$ &  & ${\sm}\bf{0.31}\,{|}\,\bf{0.59}$ & ${\sm}\bf{0.02}\,{\pm}\,\bf{0.12}$ \\	
\noalign{\smallskip} \hline \noalign{\smallskip}
\multirow{2}{*}{$\bf{NSD_{2}}$} & Marching squares &  & ${\sm}\bf{4.67}\,{|}\,\bf{2.36}$ & $0.22\,{\pm}\,1.08$ &  & ${\sm}\bf{3.30}\,{|}\,\bf{1.06}$ & ${\sm}\bf{0.19}\,{\pm}\,\bf{0.70}$ &  & ${\sm}\bf{0.41}\,{|}\,\bf{0.61}$ & ${\sm}\bf{0.00}\,{\pm}\,\bf{0.19}$ \\	
& Discrete flying edges &  & ${\sm}\bf{4.67}\,{|}\,\bf{2.36}$ & $\bf{0.21}\,{\pm}\,\bf{1.07}$ &  & ${\sm}\bf{3.30}\,{|}\,\bf{1.06}$ & ${\sm}\bf{0.19}\,{\pm}\,\bf{0.70}$ &  & ${\sm}\bf{0.41}\,{|}\,\bf{0.61}$ & ${\sm}\bf{0.00}\,{\pm}\,\bf{0.19}$ \\	
\noalign{\smallskip} \hline \noalign{\smallskip}
\multirow{2}{*}{$\bf{BIoU_{2}}$} & Marching squares &  & $\bf{0.00}\,{|}\,\bf{0.00}$ & $\bf{0.00}\,{\pm}\,\bf{0.00}$ &  & $\bf{0.00}\,{|}\,\bf{0.00}$ & $\bf{0.00}\,{\pm}\,\bf{0.00}$ &  & $\bf{0.00}\,{|}\,\bf{0.00}$ & $\bf{0.00}\,{\pm}\,\bf{0.00}$ \\	
& Discrete flying edges &  & $\bf{0.00}\,{|}\,\bf{0.00}$ & $\bf{0.00}\,{\pm}\,\bf{0.00}$ &  & $\bf{0.00}\,{|}\,\bf{0.00}$ & $\bf{0.00}\,{\pm}\,\bf{0.00}$ &  & $\bf{0.00}\,{|}\,\bf{0.00}$ & $\bf{0.00}\,{\pm}\,\bf{0.00}$ \\	
\noalign{\smallskip} \hline \noalign{\bigskip}
\multicolumn{2}{l}{\textbf{3D dataset:} HaN-Seg, voxel size:} & & 
\multicolumn{2}{c}{{(1.0, 1.0, 1.0) mm}} & &
\multicolumn{2}{c}{{(2.0, 2.0, 2.0) mm}} & &
\multicolumn{2}{c}{{(0.5, 0.5, 2.0) mm}} \\
\noalign{\smallskip} \hline \noalign{\smallskip}
\multirow{2}{*}{$\bf{HD}$} & Marching cubes &  & ${\sm}\bf{0.33}\,{|}\,\bf{0.59}$ & $\bf{0.11}\,{\pm}\,\bf{0.13}$ &  & ${\sm}\bf{0.67}\,{|}\,\bf{0.93}$ & $\bf{0.14}\,{\pm}\,\bf{0.29}$ &  & ${\sm}\bf{0.33}\,{|}\,\bf{0.44}$ & $\bf{0.05}\,{\pm}\,\bf{0.08}$ \\	
& Discrete marching cubes &  & ${\sm}\bf{0.33}\,{|}\,\bf{0.59}$ & $\bf{0.11}\,{\pm}\,\bf{0.13}$ &  & ${\sm}\bf{0.67}\,{|}\,\bf{0.93}$ & $\bf{0.14}\,{\pm}\,\bf{0.29}$ &  & ${\sm}\bf{0.33}\,{|}\,\bf{0.44}$ & $\bf{0.05}\,{\pm}\,\bf{0.08}$ \\	
\noalign{\smallskip} \hline \noalign{\smallskip}
\multirow{2}{*}{$\bf{HD_{95}}$} & Marching cubes &  & ${\sm}\bf{0.45}\,{|}\,\bf{0.99}$ & $\bf{0.17}\,{\pm}\,\bf{0.16}$ &  & ${\sm}2.85\,{|}\,\bf{1.33}$ & $0.17\,{\pm}\,0.31$ &  & ${\sm}\bf{0.56}\,{|}\,\bf{1.02}$ & $\bf{0.15}\,{\pm}\,\bf{0.15}$ \\	
& Discrete marching cubes &  & ${\sm}\bf{0.45}\,{|}\,\bf{0.99}$ & $\bf{0.17}\,{\pm}\,\bf{0.16}$ &  & ${\sm}\bf{2.83}\,{|}\,\bf{1.33}$ & $\bf{0.16}\,{\pm}\,\bf{0.32}$ &  & ${\sm}\bf{0.56}\,{|}\,\bf{1.02}$ & $\bf{0.15}\,{\pm}\,\bf{0.15}$ \\	
\noalign{\smallskip} \hline \noalign{\smallskip}
\multirow{2}{*}{$\bf{MASD}$} & Marching cubes &  & ${\sm}\bf{0.13}\,{|}\,\bf{0.72}$ & $\bf{0.13}\,{\pm}\,\bf{0.06}$ &  & ${\sm}\bf{0.52}\,{|}\,\bf{1.19}$ & $\bf{0.16}\,{\pm}\,\bf{0.10}$ &  & ${\sm}\bf{0.03}\,{|}\,\bf{0.41}$ & $\bf{0.08}\,{\pm}\,\bf{0.04}$ \\	
& Discrete marching cubes &  & ${\sm}\bf{0.13}\,{|}\,\bf{0.72}$ & $\bf{0.13}\,{\pm}\,\bf{0.06}$ &  & ${\sm}\bf{0.52}\,{|}\,\bf{1.19}$ & $\bf{0.16}\,{\pm}\,\bf{0.10}$ &  & ${\sm}\bf{0.03}\,{|}\,\bf{0.41}$ & $\bf{0.08}\,{\pm}\,\bf{0.04}$ \\	
\noalign{\smallskip} \hline \noalign{\smallskip}
\multirow{2}{*}{$\bf{ASSD}$} & Marching cubes &  & ${\sm}\bf{0.21}\,{|}\,\bf{1.31}$ & $\bf{0.13}\,{\pm}\,\bf{0.07}$ &  & ${\sm}\bf{0.59}\,{|}\,\bf{1.97}$ & $\bf{0.16}\,{\pm}\,\bf{0.12}$ &  & ${\sm}\bf{0.22}\,{|}\,\bf{0.72}$ & $\bf{0.08}\,{\pm}\,\bf{0.05}$ \\	
& Discrete marching cubes &  & ${\sm}\bf{0.21}\,{|}\,\bf{1.31}$ & $\bf{0.13}\,{\pm}\,\bf{0.07}$ &  & ${\sm}\bf{0.59}\,{|}\,\bf{1.97}$ & $\bf{0.16}\,{\pm}\,\bf{0.12}$ &  & ${\sm}\bf{0.22}\,{|}\,\bf{0.72}$ & $\bf{0.08}\,{\pm}\,\bf{0.05}$ \\	
\noalign{\smallskip} \hline \noalign{\smallskip}
\multirow{2}{*}{$\bf{NSD_{2}}$} & Marching cubes &  & ${\sm}\bf{12.4}\,{|}\,1.40$ & ${\sm}\bf{2.10}\,{\pm}\,\bf{1.61}$ &  & ${\sm}\bf{15.9}\,{|}\,\bf{1.70}$ & ${\sm}\bf{2.68}\,{\pm}\,\bf{2.30}$ &  & ${\sm}\bf{5.79}\,{|}\,\bf{1.13}$ & ${\sm}\bf{1.32}\,{\pm}\,\bf{1.00}$ \\	
& Discrete marching cubes &  & ${\sm}\bf{12.4}\,{|}\,\bf{1.30}$ & ${\sm}\bf{2.10}\,{\pm}\,\bf{1.61}$ &  & ${\sm}\bf{15.9}\,{|}\,1.90$ & ${\sm}\bf{2.68}\,{\pm}\,\bf{2.31}$ &  & ${\sm}\bf{5.79}\,{|}\,\bf{1.13}$ & ${\sm}\bf{1.32}\,{\pm}\,\bf{1.01}$ \\	
\noalign{\smallskip} \hline \noalign{\smallskip}
\multirow{2}{*}{$\bf{BIoU_{2}}$} & Marching cubes &  & ${\sm}\bf{3.69}\,{|}\,\bf{0.50}$ & ${\sm}\bf{1.25}\,{\pm}\,\bf{0.82}$ &  & ${\sm}\bf{14.7}\,{|}\,\bf{2.18}$ & ${\sm}5.89\,{\pm}\,3.85$ &  & ${\sm}\bf{2.36}\,{|}\,\bf{0.32}$ & ${\sm}\bf{0.87}\,{\pm}\,\bf{0.54}$ \\	
& Discrete marching cubes &  & ${\sm}\bf{3.69}\,{|}\,\bf{0.50}$ & ${\sm}\bf{1.25}\,{\pm}\,\bf{0.82}$ &  & ${\sm}\bf{14.7}\,{|}\,\bf{2.18}$ & ${\sm}\bf{5.87}\,{\pm}\,\bf{3.84}$ &  & ${\sm}\bf{2.36}\,{|}\,\bf{0.32}$ & ${\sm}\bf{0.87}\,{\pm}\,\bf{0.54}$ \\	
\noalign{\smallskip} \hline
\end{tabularx}
\end{small}
    \vspace{-3ex}
\end{table}
\end{document}